\documentclass[10pt,twocolumn,letterpaper]{article}
\usepackage{multirow}
\usepackage[pagenumbers]{pstyle}

\definecolor{linkblue}{rgb}{0.21,0.49,0.74}
\usepackage[pagebackref,breaklinks,colorlinks,allcolors=linkblue]{hyperref}

\title{Know-Your-Scene (KYS)-SLAM: Hierarchical Semantic-Motion Priors for Feature Matching in Stereo Visual SLAM}

\author{Preeti Chatterjee \quad Jin Lu \quad Jin Sun \quad Suchendra M. Bhandarkar\\
School of Computing, University of Georgia\\
Athens, GA 30602, USA\\
{\tt\small preeti.chatterjee@uga.edu}
}

\begin{document}
\maketitle
\raggedbottom
\begin{abstract}
Stereo visual SLAM systems built on local descriptors suffer from semantic
ambiguity, instance-level confusion, and independently moving objects, each
corrupting data association and accumulating as trajectory drift. Prevailing
semantic and dynamic SLAM methods address this through binary feature
rejection, sacrificing correspondence density and geometric support for outlier
suppression. We contend that contextual implausibility is better expressed as a
graded quantity than an exclusion criterion. We present Know-Your-Scene
(KYS)-SLAM, a modular extension of ORB-SLAM3 that supplants feature rejection
with continuous correspondence modulation. The contribution is the reframing of
contextual evidence as correspondence cost, applied entirely within feature
matching and leaving the geometric backend unmodified. Each keypoint is
augmented with semantic, panoptic, and motion priors fused through a
hierarchical compatibility formulation, in which semantic class and instance
identity enforce structural plausibility while a zero-shot motion score
down-weights features on independently moving objects. That score comes from a
training-free module fitting a depth-aware ego-motion model to background
optical flow and classifying panoptic segments via self-calibrating,
coverage-aware thresholds, so only segments with sufficient motion evidence are
penalized and static structure is left unpenalized. Penalizing correspondences
in proportion to their implausibility rather than discarding them preserves the
geometric support bundle adjustment depends on. Under one fixed configuration,
no coefficient retuned per sequence or dataset, KYS-SLAM reduces per-sequence
ATE RMSE by $17.4\%$ on outdoor KITTI and $27.7\%$ on indoor EuRoC across 21
stereo sequences with no regressions, and by 6.6\% on dynamic subsets of KITTI
Tracking and 17.8\%, up to 31.2\%, on Virtual KITTI 2 --- cross-domain transfer
across outdoor driving, indoor flight, and synthetic imagery under one set of
constants.
\end{abstract}
\vspace {-0.2in}
\section{Introduction}
\label{sec:intro}

Simultaneous Localization and Mapping (SLAM) is foundational to autonomous perception. Classical feature-based systems such as the ORB (Oriented FAST and Rotated BRIEF)-SLAM family~\cite{mur2015orb,mur2017orb,campos2021orb} achieve strong accuracy by exploiting local image features and geometric optimization, but are designed around predominantly static-scene assumptions~\cite{cadena2016past} and rely primarily on appearance-based descriptors for data association. This leaves them vulnerable to semantic ambiguity between structurally dissimilar regions, instance-level confusion among same-class objects, and corruption from independently moving entities, all of which degrade data association and propagate as trajectory drift through bundle adjustment.

Prior semantic and dynamic SLAM systems typically treat contextual evidence as a filtering signal rather than as a graded correspondence prior. Semantic-region and mask-based methods suppress observations on undesired or potentially dynamic regions, whether via semantic masks~\cite{semanticI, semanticII}, instance segmentation with multi-view geometry~\cite{dynaslam}, or optical-flow tracking~\cite{dymroslam}. Even systems that estimate graded region-level motion probabilities, such as RSO-SLAM~\cite{rsoslam}, still apply them for dynamic-region detection and feature selection rather than continuous correspondence-level modulation. More broadly, detector-driven methods inherit the generalization limits of their trained classes and can fail on unlabeled moving objects~\cite{panopticslam}, while geometric filtering depends on thresholds whose behavior varies across scenes. Such suppression-based strategies reduce the correspondence density that bundle adjustment depends on for stable optimization, and can discard static, geometrically informative features as segmentation or motion-estimation noise. We argue, instead, that contextual priors should modulate the \textit{cost} of a correspondence rather than determine its \textit{existence}.

We propose Know-Your-Scene (KYS)-SLAM, a modular extension of ORB-SLAM3 that
embeds semantic, panoptic, and motion priors into stereo matching through
hierarchical correspondence compatibility. Semantic class, temporally
consistent instance identity, and a depth-compensated scale proxy rescale the
native descriptor distance across same-instance, cross-instance, and
cross-class cases, penalizing implausible correspondences proportionally rather
than discarding them. A zero-shot motion module, requiring no training or
class-level assumptions, fits a depth-aware ego-motion model to background
optical flow and gates panoptic segments through self-calibrating thresholds,
so that only segments with sufficient motion evidence are penalized. Where
semantic and dynamic SLAM methods are typically evaluated on either outdoor
driving or indoor RGB-D benchmarks, we evaluate within a unified pipeline and
configuration on outdoor stereo KITTI~\cite{geiger2012kitti} and indoor stereo
EuRoC~\cite{mav}, excluding V203, whose stereo streams are not frame-count
matched in our preprocessing --- breaking synchronized stereo processing unless
frames are manually pruned --- and which is reported as challenging for stereo
tracking~\cite{zheng2018trifo}. KYS-SLAM achieves $17.4\%$ and $27.7\%$ average per-sequence ATE reduction over
ORB-SLAM3 across 11 KITTI and 10 EuRoC sequences without regression, and
6.6\% on dynamic subsets of KITTI Tracking~\cite{geiger2012kitti} and 17.8\%, up to 31.2\%, on Virtual KITTI 2~\cite{cabon2020vkitti2}, with every
coefficient fixed once across all datasets.

\noindent \textbf{(1)} A unified correspondence modulation framework
integrating semantic, instance-level, and motion priors into ORB-SLAM3,
replacing binary feature rejection with hierarchical cost scaling that
preserves geometric support and introduces no absolute descriptor-scale
constants.

\noindent \textbf{(2)} A zero-shot motion detection pipeline with depth-aware
ego-motion fitting, spatially local residual correction, and coverage-aware
self-calibrating thresholds, yielding bounded detection-gated scores that drive
gains on dynamic sequences while remaining largely inactive on static ones.

\noindent \textbf{(3)} Cross-domain evaluation across 21 KITTI and EuRoC sequences and
dynamic subsets of KITTI Tracking and Virtual KITTI~2, reporting per-sequence
ATE RMSE over three runs under SE(3) alignment~\cite{grupp2017evo}, with
runtime profiling and controlled robustness studies under degraded segmentation
and optical flow.
\section{Related Work}

\noindent\textbf{Optimization-Centric and Learned Visual SLAM.} Modern visual
odometry and SLAM span end-to-end methods that regress camera motion directly
from image sequences~\cite{wang2017deepvo}, hybrid systems fusing CNN-predicted
depth with direct monocular SLAM~\cite{tateno2017cnn}, and learned formulations
integrating optimization more deeply, whether over photometric, reprojection,
and inertial cues~\cite{dvislam} or through a differentiable dense
bundle-adjustment layer~\cite{droidslam}. Despite these advances,
optimization-centric geometric systems remain widely used for reliable pose
estimation: the ORB-SLAM family~\cite{mur2015orb,mur2017orb,campos2021orb}
extracts sparse ORB features for bundle adjustment, VINS-Fusion~\cite{vinsfusion}
formulates multi-sensor odometry as nonlinear factor optimization, and
SVO~\cite{svo} combines direct image alignment with sparse geometric estimation.
While geometrically strong, these systems do not use semantic labels, instance
identities, and motion priors as graded correspondence-level priors for
modulating matching costs, limiting robustness in dynamic and semantically
ambiguous scenes.

\noindent\textbf{Semantic and Dynamic SLAM.}
Semantic-region approaches remove observations from undesired or potentially
dynamic areas~\cite{semanticI}, and outdoor dense-mapping systems fuse semantic
cues, depth, and multi-view geometry to suppress moving
objects~\cite{semanticII}. DynaSLAM~\cite{dynaslam} pairs Mask~R-CNN with
multi-view geometry to detect and remove dynamic regions before inpainting the
background, DS-SLAM~\cite{dsslam} filters them by semantic segmentation and
moving-consistency checks, and DYMRO-SLAM~\cite{dymroslam} combines
Mask~R-CNN with optical-flow tracking for stereo SLAM in dynamic scenes.
RSO-SLAM~\cite{rsoslam} instead estimates region-level motion probabilities
from instance segmentation and optical flow. Panoptic formulations follow:
Panoptic-SLAM~\cite{panopticslam} couples panoptic labels with epipolar
geometry to handle unknown dynamic objects, and PSMD-SLAM~\cite{psmdslam}
embeds panoptic segmentation in multi-sensor dynamic object removal. The common
design is suppression --- observations judged \textit{dynamic} or
\textit{unreliable} are masked or removed before tracking and mapping --- which
improves robustness but treats contextual evidence as a binary decision rather
than as uncertainty in the correspondence cost. KYS-SLAM instead treats
semantic, structural, and motion evidence as continuous correspondence priors,
preserving geometric support while penalizing implausible associations.

\noindent\textbf{Optical Flow and Motion Modeling.}
Optical-flow-driven dynamic detection spans residual-based segmentation,
semantic-flow fusion, and object-level motion modeling.
FlowFusion~\cite{flowfusion} uses optical-flow residuals to highlight dynamic
regions in RGB-D point clouds within dense direct SLAM.
VDO-SLAM~\cite{vdoslam} estimates camera motion jointly with the full SE(3)
motion of tracked dynamic rigid objects, unifying static and dynamic structure.
Chen et al.~\cite{chen2022monocular} combine semantic information with
optical-flow and epipolar constraints to filter dynamic feature points in
indoor monocular SLAM, and Huang et al.~\cite{huang2024zeroshot} fuse geometric
motion models --- including a refined Longuet-Higgins
formulation~\cite{longuethiggins} --- over object proposals by multi-view
spectral clustering for zero-shot monocular motion segmentation. These methods
establish the value of flow and geometric residuals for identifying dynamic
content, but use motion reasoning for segmentation, object-motion estimation,
or feature filtering. KYS-SLAM instead converts motion evidence into a
detection-gated continuous correspondence prior within sparse feature matching,
integrated with semantic and instance-level compatibility in a hierarchical
modulation framework.

\vspace{-3mm}
\section{Know-Your-Scene (KYS)-SLAM}
KYS-SLAM augments the stereo feature-matching stage of ORB-SLAM3 with
contextual priors, leaving the geometric backbone unchanged so the gains are
attributable to the matching cost alone. Each ORB keypoint carries a prior
tuple---semantic class label, instance identity, depth-compensated scale, and
motion score---computed offline and propagated from frames to keyframes and map
points, so tracking, local mapping, and loop closing all use the original
descriptor and its priors. Rather than rejecting candidates on dynamic or
semantically inconsistent regions, it rescales each candidate's descriptor
distance by contextual plausibility, preserving geometric support while
down-weighting unreliable associations. We develop this in stages: semantic
compatibility establishes, as proof of concept, that class-level coherence
regularizes matching without overriding geometry (Sec.~\ref{sec:semantic});
panoptic encoding extends it into a structural backbone with instance identity
and depth-aware scale (Sec.~\ref{sec:panoptic}); a zero-shot motion module
converts optical-flow residuals into a per-segment motion score
(Sec.~\ref{sec:motion}); and a unified rule fuses all priors into the single
correspondence cost that defines KYS-SLAM (Sec.~\ref{sec:unified}).

\vspace{-2mm}
\subsection{From Semantic Labels to Correspondence Cost}
\label{sec:semantic}
\vspace{-1mm}
We first establish the core mechanism of KYS-SLAM in its simplest form; rather than accepting or rejecting a correspondence, we reweight its matching cost by contextual compatibility. As proof of concept for this cost-modulation principle, the semantic class of each keypoint serves as an elementary compatibility measure, further extended by the panoptic and motion stages.

\begin{figure*}[t]
  \centering
  \begin{minipage}{0.38\linewidth}\centering\includegraphics[width=\linewidth]{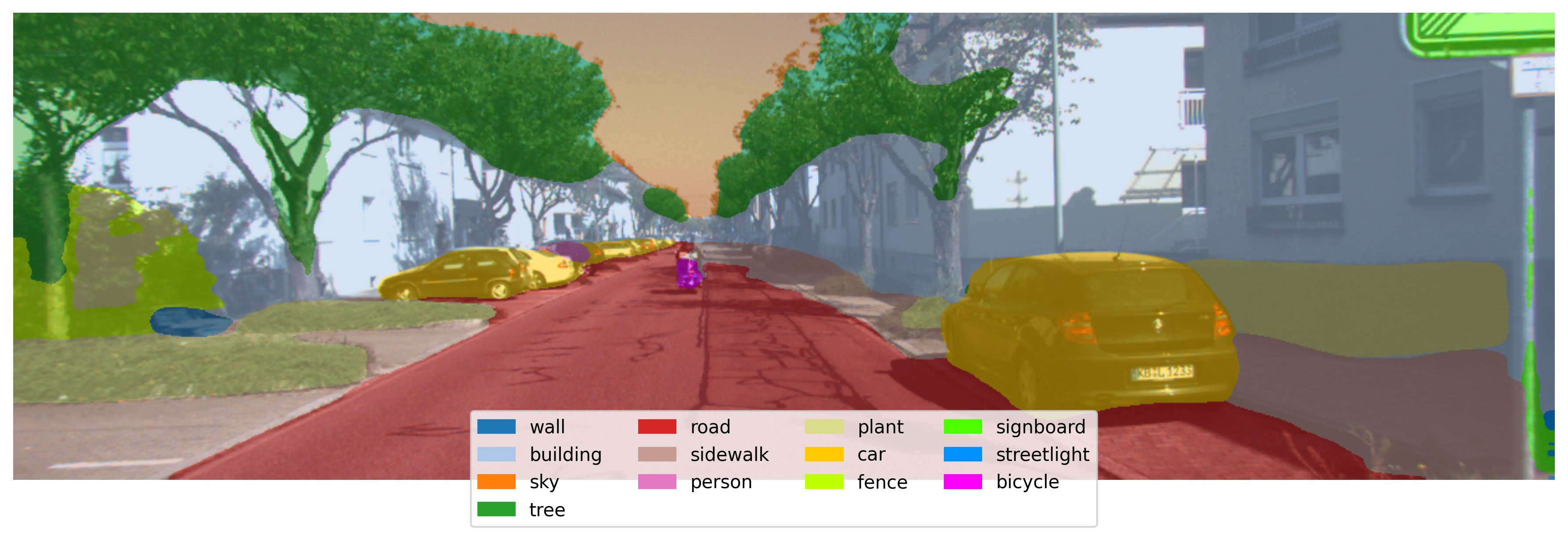}\end{minipage}\hfill
  \begin{minipage}{0.58\linewidth}\centering\includegraphics[width=\linewidth]{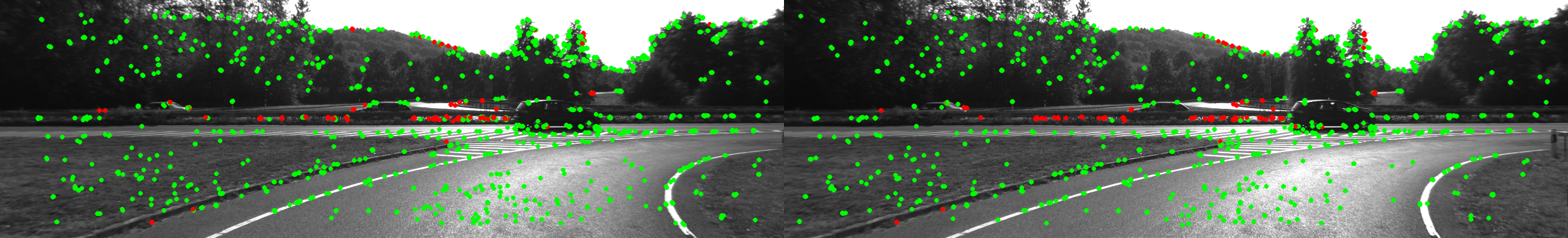}\end{minipage}
  \caption{Class-level semantic matching on KITTI. Left: ADE20K Mask2Former overlay on a crowded Seq.~00 frame. Right: stereo correspondences on a Seq.~01 frame with moving vehicles, where \emph{green} marks class-consistent matches and \emph{red} class mismatches.}
  \label{fig:sem_overlay}
\end{figure*}

Dense semantic maps $\mathcal{S}_t$ are precomputed offline with Mask2Former~\cite{cheng2022mask2former} on a Swin backbone~\cite{liu2021Swin} fine-tuned on ADE20K~\cite{zhou2017scene,zhou2019semantic}, assigning each pixel one of $C\!=\!150$ labels covering the dominant classes in KITTI and EuRoC (Fig.~\ref{fig:sem_overlay}, left). Each keypoint $i$ inherits a label $\ell_i\!\in\!\{1,\dots,C\}$ by indexing $\mathcal{S}_t$ at its rounded ORB coordinate.

The simplest variant, \emph{Hard Mismatch Rejection}, discards any candidate with $\ell_i\!\neq\!\ell_j$ and is sensitive to segmentation noise at boundaries, where a single-pixel flip discards a valid match and thins the correspondence density. We therefore treat agreement as a graded entity: each class name is embedded as a FastText vector~\cite{bojanowski2017enriching} of dimension $300$, sign-binarized to $\mathbf{v}_\ell\!\in\!\{0,1\}^{300}$, from which a symmetric inter-class matrix $\mathbf{S}\!\in\!\mathbb{R}^{C\times C}$ is precomputed once,
\vspace{-1mm}
\begin{equation}
\mathbf{S}(\ell_i,\ell_j) = \bigl\| \mathbf{v}_{\ell_i} \oplus \mathbf{v}_{\ell_j} \bigr\|_1 ,
\label{eq:semantic_distance}
\end{equation}

\noindent with $\oplus$ the bitwise exclusive-or and $\|\cdot\|_1$ the count of differing bits (i.e., Hamming distance). The values span $[0,169]$ over the $C\!=\!150$ categories, with close classes (\textit{road}/\textit{sidewalk}) yielding small $\mathbf{S}$ values and distant ones (\textit{road}/\textit{sky}) large $\mathbf{S}$ values. Implemented as a fixed lookup, the cue costs one table access per match (Fig.~\ref{fig:sem_overlay}, right).

\emph{Soft Filtering} retains a candidate only when $\mathbf{S}(\ell_i,\ell_j)\!<\!\tau_{\mathrm{sem}}$ ($\tau_{\mathrm{sem}}\!=\!110$). Continuous \emph{fusion} embodies the modulation principle: for a candidate $(i,j)$ with raw ORB Hamming distance $d\!\in\![0,256]$,
\vspace{-1mm}
\begin{equation}
d_{\mathrm{sem}}(i,j) = \alpha\, d + (1-\alpha)\, \mathbf{S}(\ell_i,\ell_j),\quad \alpha\!\in\![0,1],
\label{eq:kys_sem}
\end{equation}
where the two Hamming distance terms $d$ and $\mathbf{S}$ share comparable ranges and are fused without normalization while the geometric descriptor $d$ stays dominant. This resolves the boundary sensitivity of \emph{Hard Mismatch Rejection}; i.e., a close-class flip incurs small $\mathbf{S}$ value and survives, whereas cross-class boundaries yield large $\mathbf{S}$ values and are down-weighted, not rejected. Sweeping $\alpha$ on Seq.~02, 07 and 08 (Supp.~A), geometry-dominant settings ($\alpha\!\ge\!0.6$) keep ATE within a narrow band of the best value, while strongly semantic-dominant ones degrade sharply, reaching $10.40$\,m on Seq.~02 at $\alpha\!=\!0.1$ against $4.66$\,m at $\alpha\!=\!0.7$. Every coefficient in KYS-SLAM is a global constant, set once and applied unchanged to all $21$ sequences of both datasets: $\alpha\!=\!0.7$ is a representative value in this stable range, not a per-sequence optimum. Fixed global constants of this kind are standard in feature-based SLAM; KYS-SLAM additionally learns no parameters. Class identity alone, however, cannot disambiguate multiple objects of one category nor reason about depth; two vehicles can share a class label yet differ structurally, motivating the panoptic backbone, which augments the formulation with instance identity and depth-aware scale.

\begin{figure*}[t]
  \centering
  \begin{minipage}{0.38\linewidth}
    \centering
    \includegraphics[width=\linewidth]{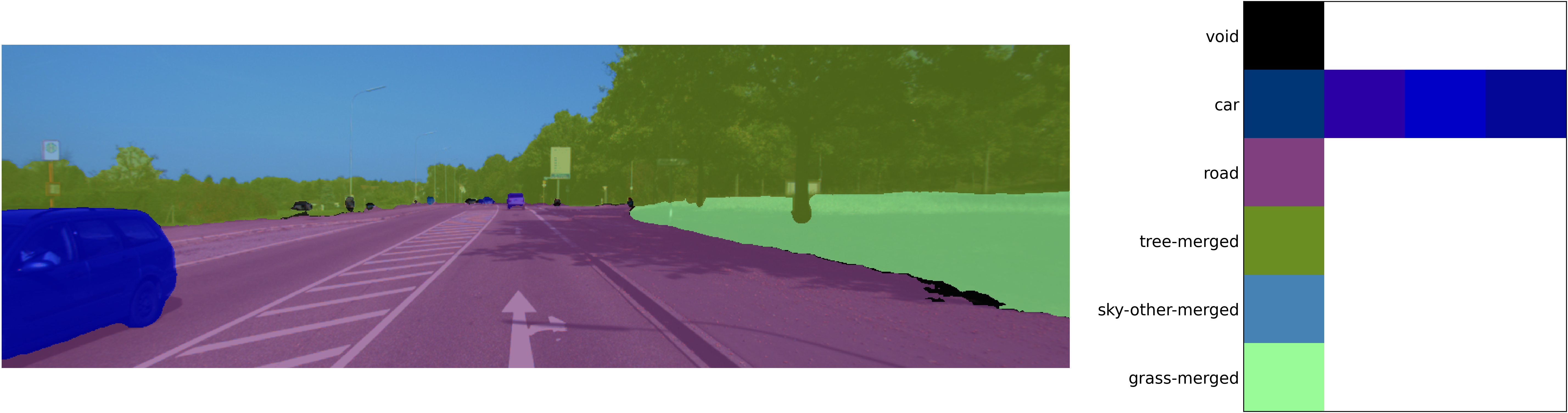}
  \end{minipage}
  \hfill
  \begin{minipage}{0.58\linewidth}
    \centering
    \includegraphics[width=\linewidth]{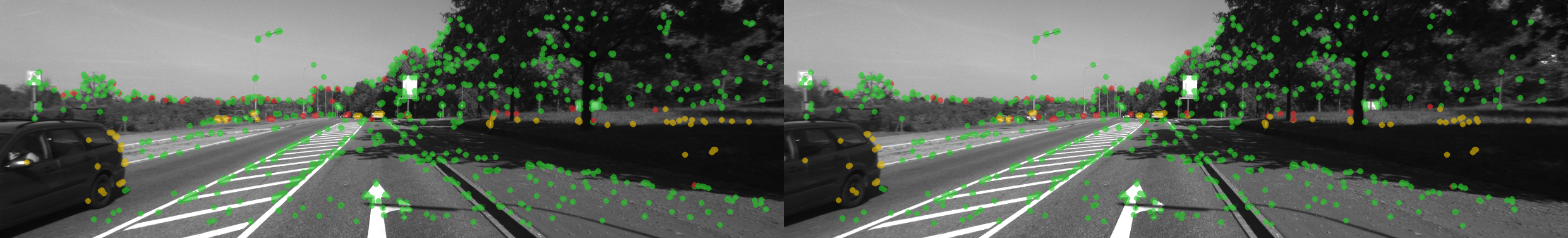}
  \end{minipage}
  \caption{Left: Panoptic segmentation with temporally consistent instance IDs. Right: Stereo correspondences under hierarchical panoptic compatibility. Green denotes same-instance matches, yellow cross-instance matches within the same class, and red cross-class mismatches.}
  \label{fig:panoptic_match_stacked}
\end{figure*}

\subsection{Panoptic Structural Backbone}
\label{sec:panoptic}
\vspace{-1mm}
Class labels alone cannot resolve intra-class ambiguity or enforce geometric plausibility across depth. We therefore extend the cue of Sec.~\ref{sec:semantic} to \emph{instance-aware} reasoning over semantic compatibility, instance identity, and coarse 3D scale. This panoptic encoding forms the structural backbone which the motion module (Sec.~\ref{sec:motion}) and the unified rule (Sec.~\ref{sec:unified}) extend. Per-pixel semantic labels and instance masks are obtained offline with KMaX-DeepLab~\cite{kmaxdeeplab} (ResNet-50~\cite{he2016deep}, COCO~\cite{cocodataset}), as implemented in the DeepLab2 library~\cite{yu2023kmax}, giving a semantic map $\ell(u,v)$ and instance map $s(u,v)$ for each image $I_t$ (Fig.~\ref{fig:panoptic_match_stacked}, left).

\noindent \textbf{Temporal instance association.}
Since $s(u,v)$ is frame-local, instances are linked over time by centroid tracking: for each active track $\tau$ and current-frame detection $\delta$ we compute the Euclidean distance between mask centroids and solve the assignment with the Hungarian algorithm~\cite{Kuhn1955TheHM}. Matches above $d_{\max}=100$\,px are rejected and tracks unmatched for more than $5$ consecutive frames dropped, yielding identities that persist across frames. Photometric alternatives such as DOT~\cite{Ballester2021DOT} track objects to estimate their motion; we need identity only, deferring motion to Sec.~\ref{sec:motion}.

\noindent \textbf{Scale-aware structural encoding.}
To separate same-class instances at different depths, each segment carries a depth-compensated scale proxy. Depth follows from disparity $\Delta(u,v)$ as $Z(u,v)=fB/\Delta(u,v)$, with $f$ the focal length and $B$ the baseline. For the $k$-th panoptic segment $\mathcal{S}_k$ with pixel area $|\mathcal{S}_k|$,
\vspace{-1mm}
\begin{equation}
\bar Z_k=\frac{1}{|\mathcal{S}_k|}\sum_{(u,v)\in\mathcal{S}_k} Z(u,v),
\qquad
a_k \propto |\mathcal{S}_k|\,\bar Z_k^{2}.
\label{eq:panoptic_scale}
\end{equation}
Under the pinhole camera model projected area scales as $1/Z^2$, so $a_k$ compensates for depth and yields a scale measure comparable across ranges. Each keypoint thus inherits the panoptic attributes $(\ell_i, \theta_i, a_i)$, where $\theta_i$ is its instance ID and $a_i$ its segment scale.

\noindent \textbf{Hierarchical compatibility.}
Let $d$ be the ORB Hamming distance for a candidate pair $(A,B)$, and let $\mathbf{S}(\ell_A,\ell_B)$ be the semantic distance of Eq.~\eqref{eq:semantic_distance} (a $133\times133$ COCO table, range $[0,172]$, scaled by $S_{\max}=155$, the typical maximum, so the term exceeds $1$ only for rare highly dissimilar pairs). We define a bounded scale discrepancy
\vspace{-1mm}
\begin{equation}
\Delta a_{AB}=\frac{|a_A-a_B|}{\max(a_A,a_B)}\in[0,1].
\label{eq:area_penalty}
\end{equation}

\noindent Since panoptic segmentation is run independently on each image, the
same physical object may receive different labels or instance identities across
views, while its area remains approximately preserved. The area discrepancy
$\Delta a_{AB}$ acts as a geometrically stable cue that backstops the noisier
semantic and instance cues: it stays small for true matches and prevents
per-view label or instance disagreement from inflating their structural
multiplier, while genuinely scale-inconsistent pairs are penalized more
strongly. Identity is likewise assigned within a stream, so the same-instance
tier governs temporal correspondences while $\ell$ and $\Delta a$ carry the
stereo case; the same holds on loop closure, where tracks have long expired.
The structural compatibility multiplicatively modulates $d$ in three tiers,
preserving its native Hamming scale:
\vspace{-1mm}
\begin{equation}
\Phi_{\mathrm{pan}}=
d\cdot
\begin{cases}
\eta_1+\eta_2\dfrac{\mathbf{S}(\ell_A,\ell_B)}{S_{\max}}+\eta_3\,\Delta a_{AB}, & \ell_A\neq \ell_B,\\[4pt]
\gamma+\eta_3\,\Delta a_{AB}, & \ell_A=\ell_B,\ \theta_A\neq \theta_B,\\[2pt]
\gamma, & \ell_A=\ell_B,\ \theta_A=\theta_B,
\end{cases}
\label{eq:panoptic_cost}
\end{equation}
\noindent with $(\eta_1,\eta_2,\eta_3)=(0.70,0.20,0.10)$ and $\gamma=0.75$.
Same-instance pairs receive the discount $\gamma$; cross-instance same-class
pairs are tempered upward by scale discrepancy; and cross-class pairs receive a
discount governed jointly by the semantic and scale terms, weakest for distant
classes and strongest for proximate ones. The discounts stay gentle by design,
down-weighting implausible pairs without rejecting them. The tiers are
therefore not globally ordered at fixed descriptor distance: since
$\eta_1<\gamma$, a semantically proximate cross-class pair may receive a
discount at or slightly below the same-instance value. This is intended, as the
ordering is by contextual implausibility rather than tier membership, and a
near-class pair such as road/sidewalk is not less plausible than one within a
single instance. These coefficients are fixed once across all sequences
(Sec.~\ref{sec:semantic}), and accuracy is stable over a broad range about each
(Supp.~A). Fig.~\ref{fig:panoptic_match_stacked} (right) illustrates the three
tiers: same-instance (green), cross-instance same-class (yellow), and
cross-class (red). The formulation is deliberately modular: each stage consumes
only standard outputs---class labels for the semantic cue
(Sec.~\ref{sec:semantic}) and instance masks for the backbone---so any
segmenter furnishing these may be substituted without modifying the matcher,
leaving the framework free to incorporate stronger models as they emerge.

\vspace{-1mm}
\subsection{Zero-Shot Motion Detection}
\label{sec:motion}
\vspace{-1mm}
The panoptic backbone reasons about structure but not motion; a parked and a moving car are treated identically. We add a motion prior that flags independently moving objects without training or class-level assumptions. Since a moving camera induces apparent motion everywhere, we predict the flow a static scene would produce and analyze the residual (Supp. ~H). Every threshold described in the following is either a fixed constant or a multiple of a per-frame statistic; thus one configuration applies unchanged to all sequences.

\begin{figure*}[t]
  \centering
  \includegraphics[width=\linewidth,height=0.22\textheight,keepaspectratio]{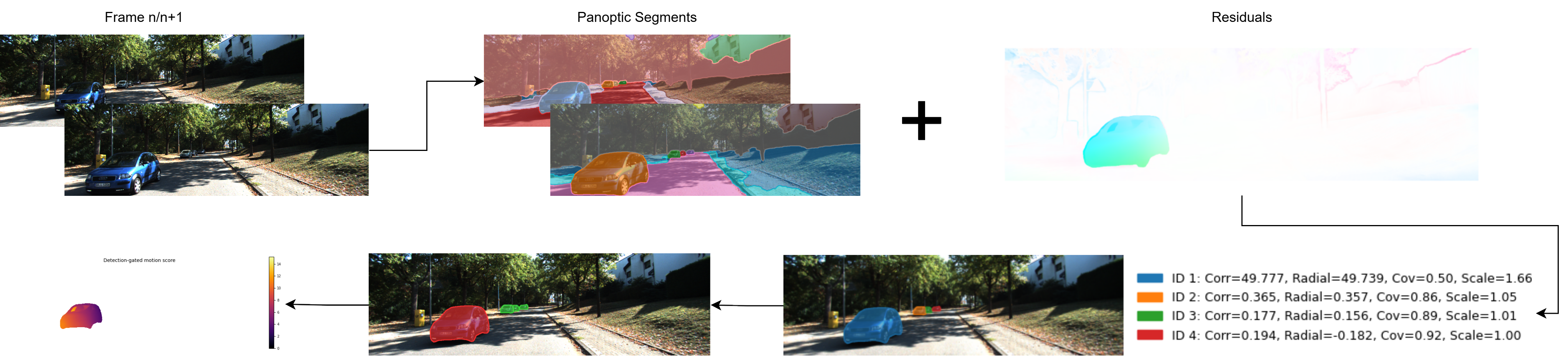}
  \caption{Motion-detection pipeline on a KITTI frame. Consecutive frames and panoptic segments yield the ego-motion residual; per \emph{thing} segment we report the corrected-residual magnitude (\textbf{Corr}), radial residual (\textbf{Radial}), background coverage (\textbf{Cov}), and coverage threshold multiplier (\textbf{Scale}). Only the moving vehicle (ID~1, Corr=49.8) exceeds the per-frame background level; the static objects (ID~2--4, Corr$<$1) do not. The final panel shows the detection-gated motion score $m_k=\min(5[\log(1+\rho_k)]^2,100)$: static segments are gated to zero while the mover receives a bounded score rather than its large raw residual.}
  \label{fig:motion_pipeline}
\end{figure*}

\noindent \textbf{Ego-motion model on background only.}
Dense flow $\mathbf{F}_{\mathrm{obs}}$ is estimated with UniMatch~\cite{xu2023unifying}, and depth $Z$ (capped at $255$) is derived from disparity $\Delta$; pixel coordinates are normalized as $x\!=\!(u-W/2)/W$, $y\!=\!(v-H/2)/H$. Under a rigid static scene, image motion follows the depth-aware linearized Longuet-Higgins model~\cite{longuethiggins,huang2024zeroshot}
\vspace{-1mm}
\begin{equation}
\begin{aligned}
x_{\mathrm{m}} &= c_1 + c_2 Z^{-1} - c_3\,x Z^{-1} - c_4\,y + c_5\,x^2 - c_6\,x y,\\
y_{\mathrm{m}} &= c_7 + c_8 Z^{-1} - c_3\,y Z^{-1} - c_4\,x + c_5\,x y + c_6\,y^2 ,
\end{aligned}
\label{eq:lh_model}
\end{equation}
with $(x_{\mathrm{m}},y_{\mathrm{m}})$ the predicted flow in the normalized image plane. We fit its eight parameters on \emph{stuff} pixels only, excluding a $20$\,px border, so moving \emph{things} are withheld by construction. The fit is doubly robust: the mask removes most dynamic content a priori, and an IRLS solve with the Huber loss~\cite{huber1964robust} ($\delta_H\!=\!1$\,px, $5$ iterations) down-weights contamination from mislabeled pixels, yielding a static-scene flow $\mathbf{F}_{\mathrm{m}}$ stable under imperfect segmentation.

\noindent \textbf{Residual extraction and drift correction.}
The raw residual $\mathbf{r}\!=\!\mathbf{F}_{\mathrm{obs}}\!-\!\mathbf{F}_{\mathrm{m}}$ is non-zero on static structure due to flow noise, depth error, and linearization, so a global threshold flags static surfaces yet misses slow motion. We subtract a spatially-local reference: \textit{stuff}-pixel residuals smoothed by a wide Gaussian ($\sigma\!=\!60$\,px) and divided by an identically smoothed coverage map form a drift field $\bar{\mathbf{r}}_{\mathrm{bg}}$, giving the corrected residual $\tilde{\mathbf{r}}\!=\!\mathbf{r}\!-\!\bar{\mathbf{r}}_{\mathrm{bg}}$; if the road around a parked car drifts, the car inherits that drift, which the subtraction cancels. We also retain the signed radial residual $\tilde r^{\parallel}\!=\!\langle\tilde{\mathbf{r}},\mathbf{F}_{\mathrm{m}}\rangle/\|\mathbf{F}_{\mathrm{m}}\|$, the projection onto the predicted-flow direction, which exposes optical-axis motion that magnitude misses.


\noindent \textbf{Dual detection with self-calibrating thresholds.}
For each \textit{thing} segment $k$ we compute the mean corrected magnitude $\bar r_k$, mean radial residual $\bar r^{\parallel}_k$, and mean coverage $c_k$ (reported per segment as Corr, Radial, and Cov in Fig.~\ref{fig:motion_pipeline}); over the \textit{stuff} segments we form three per-frame references: the mean magnitude $\bar m_{\mathrm{bg}}$, and the mean $\mu^{\parallel}$ and standard deviation $\sigma^{\parallel}$ of the radial residual, recomputed every frame so the thresholds self-calibrate. A segment with $c_k\!<\!0.05$ lacks background context and is forced \textit{static}; otherwise a coverage scale $\omega_k\!=\!\operatorname{clip}(1/(c_k\!+\!0.1),1,4)$ raises its detection thresholds by up to $4\times$ as background support weakens. A segment is deemed \textit{dynamic} if either test fires:
\vspace{-1mm}
\begin{equation}
\bar r_k > 2\,\bar m_{\mathrm{bg}}\,\omega_k
\;\;\text{or}\;\;
\bigl|\bar r^{\parallel}_k\!-\!\mu^{\parallel}\bigr| > 2\,\sigma^{\parallel}\,\omega_k\,\psi ,
\label{eq:detection}
\end{equation}

\noindent the magnitude test thresholding at twice the background mean and the radial test at a two-sigma deviation from it, with the latter gated to $c_k\!\ge\!0.3$ for well-defined direction. The guard $\psi\!=\!\operatorname{clip}(2|\mu^{\parallel}|/\sigma^{\parallel},1,5)$ handles camera rotation, which the linearized model cannot fully absorb and which therefore injects a coherent radial residual across the background, inflating $|\mu^{\parallel}|$ relative to $\sigma^{\parallel}$. As rotation grows, $\psi$ rises and widens the radial-tolerance bound so camera-induced flow is not mistaken for object motion; $\psi\!=\!1$ when rotation is negligible, and the cap at $5$ ensures even strong rotation does not fully suppress the true moving objects. A segment flagged by the radial test alone is tagged \emph{radial-only}. As every threshold is a multiple of a per-frame statistic, one configuration transfers across scenes without tuning.

\noindent \textbf{Bounded depth-robust motion score and detection gating.}
Raw motion magnitude conflates a near static object with a distant moving one, so we map the background-normalized ratio $\rho_k\!=\!\bar r_k/\bar m_{\mathrm{bg}}$ (clipped to $[0,100]$) through a log-squared transform
\vspace{-3mm}
\begin{equation}
m_k = \min\!\bigl(5\,[\log(1\!+\!\rho_k)]^2,\; 100\bigr),
\label{eq:logsq}
\end{equation}
collapsing static background to $m_k\!\lesssim\!2$ while lifting genuine motion above $20$: the logarithm compresses the tail, the square restores the contrast it flattens, and the constant only sets the scale the gates of Sec.~\ref{sec:unified} are read on, both falling on a plateau (Supp.~F). The clip at $100$ makes the signal scale-invariant, so any fixed threshold behaves identically across sequences. The score is detection-gated: the dual test of Eq.~(\ref{eq:detection}), not the raw residual, decides what exports. A parked vehicle with depth noise and a slow mover may carry similar residuals, yet only the latter propagates. An unflagged segment exports nothing; a flagged one exports a confidence-weighted residual vector---the background-normalized corrected residual $\tilde{\mathbf{r}}/\bar m_{\mathrm{bg}}$ scaled by $m_k/100$ for magnitude-flagged segments and by unit weight for radial-only ones (whose low $m_k$ would otherwise suppress the radial signal), with its norm clipped to $100$ and taken as the exported prior $\mu_k$, inherited by each keypoint of that pixel as the scalar $\mu_i$ entering Sec.~\ref{sec:unified}.

\subsection{Unified Motion-Panoptic Matching}
\label{sec:unified}
The complete system unifies the panoptic structural backbone (Sec.~\ref{sec:panoptic}) and the zero-shot motion prior (Sec.~\ref{sec:motion}) into a single modulated descriptor distance, the modular contribution of KYS-SLAM to ORB-SLAM3. Each keypoint carries the four-channel prior $(\ell_i,\theta_i,a_i,\mu_i)$: semantic label, instance identity, depth-compensated scale, and exported motion prior. For a candidate correspondence $(A,B)$ with raw Hamming distance $d$, modulation proceeds in three stages.

\noindent \textbf{Hard motion gating.}
Two unilateral gates precede continuous modulation, removing correspondences that motion evidence renders unusable. Their thresholds differ because they apply to distinct ORB-SLAM3 pipelines with different error sensitivities. Stereo candidates feed triangulation, where motion error propagates into map-point depth via disparity, and are discarded at the conservative $\mu_i>8$. General descriptor matching, used for temporal tracking against already-triangulated points where error is bounded by reprojection, tolerates mild motion and rejects only at the extreme $\mu_i>20$---the regime where independent motion overrides any semantic, instance, or descriptor similarity. A single threshold would misfit one or the other. Both gates are unilateral because a correspondence is only as reliable as its more dynamic endpoint; keypoints without valid panoptic labels bypass modulation and retain their native distance.

\noindent \textbf{Structural and motion modulation.}
Surviving correspondences are scaled first by the three-tier structural factor $\Phi_{\mathrm{pan}}$ of Eq.~\eqref{eq:panoptic_cost}, then conditioned by the motion prior. Using the worst-case pair value $\bar\mu\!=\!\max(\mu_A,\mu_B)$, the final KYS-SLAM distance is
\setlength{\abovedisplayskip}{3pt}
\setlength{\belowdisplayskip}{3pt}
\begin{equation}
d_{\mathrm{KYS}} \;=\; \underbrace{\Phi_{\mathrm{pan}}}_{\text{structural}} \,\bigl(1 + \underbrace{\lambda_t\,\log(1+\bar\mu)}_{\text{motion}}\bigr),
\label{eq:unified}
\end{equation}
where the motion sensitivity $\lambda_t$, with $t$ denoting the structural tier, takes values $0.08$, $0.05$, and $0.02$ for cross-class, cross-instance, and same-instance pairs, respectively; accuracy is stable across a fourfold variation about these values (Supp.~B). The single expression realizes the full correspondence hierarchy: descriptor distance remains primary, modulated by semantic, instance, and scale compatibility, while motion sensitivity follows structural plausibility. A cross-class pair under motion is structurally and dynamically implausible, drawing the steepest motion penalty; within an instance, residual motion likely reflects boundary or flow noise rather than true failure, and is penalized least. The sub-linear $\log(1+\bar\mu)$ raises cost monotonically without letting one large residual dominate the descriptor, while the worst-case aggregate mirrors the gates by conditioning on the more dynamic endpoint.
Through Eq.~\eqref{eq:unified}, the entire semantic, panoptic, and motion pipeline collapses into one descriptor-distance evaluation per candidate, leaving the tracking, local mapping, and loop-closing optimization of ORB-SLAM3 entirely unchanged.
\section{Experiments and Results}
\noindent\textbf{Setup.}
We evaluate KYS-SLAM as a stereo extension of ORB-SLAM3 on KITTI Odometry~\cite{geiger2012kitti} and EuRoC MAV~\cite{mav}, spanning outdoor driving with dynamic traffic and indoor flight with little independent motion. We report ATE RMSE under $SE(3)$ Umeyama alignment with evo~\cite{grupp2017evo}, averaged over three runs, on an Intel Xeon W-2295 and NVIDIA RTX A6000. We use all 11 KITTI and 10 EuRoC sequences, excluding EuRoC V203 (Sec.~\ref{sec:intro}). Every parameter is fixed once across all sequences and both datasets. Baseline numbers are drawn from peer-reviewed publications under matched stereo evaluation; ORB-SLAM3 is the only system we re-run on our hardware. \textbf{DROID-SLAM$^\ast$} denotes the DROID-SLAM baseline retrained from
scratch for stereo under a controlled setting by DVI-SLAM~\cite{dvislam},
rather than the original model trained on synthetic monocular video (Supp.~J).

\noindent\textbf{Main Results.}
\begin{table*}[t]
\centering
\begingroup
\scriptsize
\setlength{\tabcolsep}{2.1pt}
\renewcommand{\arraystretch}{0.92}
\setlength{\abovecaptionskip}{2pt}
\setlength{\belowcaptionskip}{2pt}
\caption{KITTI Odometry ATE RMSE (m); lower is better. Avg.~\%$\downarrow$:
mean per-sequence reduction vs. ORB-SLAM3 over all sequences;
Dyn.~\%$\downarrow$: the same measure over the representative dynamic
sequences (marked $^{*}$), those with the most independent object motion.
Best bolded.}
\label{tab:kitti}
\begin{tabular*}{\textwidth}{@{\extracolsep{\fill}}l|ccccccccccc|cc}
\noalign{\vskip 1pt}\hline\noalign{\vskip 1pt}
\textbf{Method} & 00 & 01 & 02$^{*}$ & 03 & 04 & 05 & 06 & 07 & 08$^{*}$ & 09 & 10$^{*}$ &
\textbf{Avg.~\%$\downarrow$} & \textbf{Dyn.~\%$\downarrow$} \\
\noalign{\vskip 1pt}\hline\noalign{\vskip 1pt}
ORB-SLAM3  & 1.27 & 13.11 & 5.68 & 1.30 & 0.25 & 0.99 & 0.92 & 0.47 & 3.72 & 1.87 & 1.32 & 0.0 & 0.0 \\
ORB-SLAM2  & 1.27 & 10.40 & 5.70 & 0.79 & 0.30 & 0.87 & 0.74 & 0.50 & 3.58 & 3.57 & 1.11 & -0.6 & 6.4 \\
Stereo-DSO~\cite{stereoDSO}  & 1.35 & 10.64 & 5.57 & 0.85 & 0.21 & 0.86 & 0.78 & 0.50 & 3.55 & 3.31 & 1.07 & 3.1 & 8.5 \\
DynaSLAM~\cite{dynaslam}   & 1.40 & 9.40  & 6.70 & 0.60 & 0.20 & 0.80 & 0.80 & 0.50 & 3.50 & \textbf{1.60} & 1.20 & 11.7 & -1.0 \\
Semantic I~\cite{semanticI} & 1.23 & 24.50 & 5.35 & 0.79 & 0.18 & 0.77 & 0.67 & 0.48 & 3.57 & 3.58 & 0.95 & -2.1 & 12.6 \\
Semantic II~\cite{semanticII}& 1.21 & 10.00 & 4.96 & --   & 0.20 & 0.80 & \textbf{0.50} & 0.50 & 3.50 & 3.20 & 1.00 & -- & 14.3 \\
VDO-SLAM~\cite{vdoslam}   & 1.35 & 10.80 & 5.33 & \textbf{0.55} & 0.18 & 0.75 & 0.96 & 0.50 & 3.59 & 3.74 & 1.05 & 3.7 & 10.0 \\
AirDOS~\cite{airdos}     & 1.34 & 9.71  & 7.24 & 0.88 & 0.19 & \textbf{0.73} & 0.62 & 0.47 & 3.35 & 3.54 & 1.49 & 1.4 & -10.1 \\
RSO-SLAM~\cite{rsoslam}   & 1.27 & 9.80  & 5.30 & 0.56 & 0.18 & 0.75 & 0.74 & 0.50 & 3.49 & 1.70 & 1.16 & 16.5 & 8.3 \\
\noalign{\vskip 1pt}\hline\noalign{\vskip 1pt}
+Sem       & 1.18 & 12.10 & 4.66 & 1.14 & 0.21 & 0.84 & 0.97 & 0.47 & 3.42 & 1.76 & 1.06 & 9.5 & 15.2 \\
+Pano      & \textbf{0.98} & \textbf{9.35} & 4.58 & 1.14 & 0.19 & 0.81 & 0.90 & \textbf{0.40} & 3.39 & 1.75 & 0.99 & 16.6 & 17.7 \\
KYS-SLAM   & 1.16 & 11.20 & \textbf{4.46} & 1.03 & \textbf{0.17} & 0.82 & 0.87 & \textbf{0.40} & \textbf{3.12} & 1.70 & \textbf{0.91} & \textbf{17.4} & \textbf{22.9} \\
\hline
\end{tabular*}
\endgroup
\vspace{-1mm}
\end{table*}
\begin{table*}[t]
\centering
\begingroup
\scriptsize
\setlength{\tabcolsep}{2.1pt}
\renewcommand{\arraystretch}{0.92}
\setlength{\abovecaptionskip}{2pt}
\setlength{\belowcaptionskip}{2pt}
\caption{EuRoC MAV ATE RMSE (m). Lower is better. best bolded.}
\label{tab:euroc}
\begin{tabular*}{\textwidth}{@{\extracolsep{\fill}}l|cccccccccc|c}
\noalign{\vskip 1pt}\hline\noalign{\vskip 1pt}
\textbf{Method} & MH01 & MH02 & MH03 & MH04 & MH05 & V101 & V102 & V103 & V201 & V202 & \textbf{Avg} \\
\noalign{\vskip 1pt}\hline\noalign{\vskip 1pt}
ORB-SLAM3       & 0.029 & 0.019 & 0.024 & 0.085 & 0.052 & 0.035 & 0.025 & 0.061 & 0.041 & 0.028 & 0.040 \\
VINS-Fusion~\cite{vinsfusion}      & 0.540 & 0.460 & 0.330 & 0.780 & 0.500 & 0.550 & 0.230 & --    & 0.230 & 0.200 & --    \\
SVO ~\cite{svo}             & 0.040 & 0.070 & 0.270 & 0.170 & 0.120 & 0.040 & 0.040 & 0.070 & 0.050 & 0.090 & 0.096 \\
DROID-SLAM$^\ast$ & 0.065 & 0.042 & 0.098 & 0.191 & 0.133 & 0.063 & 0.045 & 0.043 & 0.040 & 0.054 & 0.077 \\
DVI-SLAM ~\cite{dvislam}        & 0.043 & 0.041 & 0.064 & 0.148 & 0.114 & 0.063 & 0.040 & \textbf{0.031} & 0.049 & 0.050 & 0.064 \\
\noalign{\vskip 1pt}\hline\noalign{\vskip 1pt}
+Sem            & 0.015 & 0.017 & \textbf{0.022} & 0.059 & 0.049 & 0.035 & 0.024 & 0.061 & 0.029 & 0.030 & 0.034 \\
+Pano           & 0.014 & \textbf{0.016} & \textbf{0.022} & \textbf{0.052} & \textbf{0.039} & 0.032 & 0.017 & 0.046 & 0.020 & 0.023 & \textbf{0.028} \\
KYS-SLAM        & \textbf{0.013} & 0.017 & \textbf{0.022} & 0.053 & 0.042 & \textbf{0.031} & \textbf{0.016} & 0.050 & \textbf{0.018} & \textbf{0.021} & \textbf{0.028} \\
\hline
\end{tabular*}
\endgroup
\vspace{-2mm}
\end{table*}

On KITTI (Tab.~\ref{tab:kitti}), the panoptic backbone cuts per-sequence ATE by
$16.6\%$ over ORB-SLAM3 and the full system by $17.4\%$, rising to $22.9\%$ on
the dynamic subset and exceeding DynaSLAM and RSO-SLAM on average. Several
baselines build on ORB-SLAM2, which beats ORB-SLAM3 on $6$ of $11$ sequences
here, yet KYS-SLAM surpasses them on average despite the harder backbone. On
EuRoC (Tab.~\ref{tab:euroc}), where independent motion is scarce, the backbone
carries the gain while the motion module stays largely inactive, isolating
structural generalization; KYS-SLAM also beats the retrained
DROID-SLAM$^\ast$ and DVI-SLAM on average despite being training-free. Under one fixed
configuration no sequence regresses below ORB-SLAM3, whereas competing methods
do on individual sequences (Semantic~I on Seq.~09, DynaSLAM on Seq.~00).

 
\noindent\textbf{Runtime and system cost.}
Measured on an NVIDIA RTX~A6000 under the protocol of Supp.~C, the modulation
adds $6.8$--$8.2$\,ms/frame to ORB-SLAM3's tracking stage on Seq.~03, 04 and 06
and $0.22$--$0.52$\,GB of peak host memory; a further $19$--$21$\,ms is
deserialisation of precomputed priors, an artefact of file-based storage rather
than of the formulation. Prior generation sums to $1696$\,ms/frame in
off-the-shelf perception models, plus $23.4$\,ms/frame of CPU-side fusion
introduced here; these stages are independent and overlap when run concurrently
(Supp.~C). The modulation is thus $\approx0.4\%$ of pipeline cost. With priors
resident, tracking runs at $32.6$--$35.7$\,ms/frame ($28$--$31$\,Hz), and
$18$--$19$\,Hz with deserialisation, so the matching stage introduced here is
real-time against KITTI's $10$\,Hz capture rate. KYS-SLAM \emph{as a whole} is
not optimised for real-time operation, as with prior semantic-SLAM systems that
precompute segmentation offline; the accuracy gains and the computational
burden are separable, the latter lying in substitutable perception modules.

\section{Analysis: Ablation and Behavior }
\vspace{-1mm}
\label{sec:analysis}
\label{sec:ablation}

\begin{table}[t]
\centering
\begingroup
\scriptsize
\setlength{\tabcolsep}{4pt}
\renewcommand{\arraystretch}{0.90}
\setlength{\abovecaptionskip}{2pt}
\setlength{\belowcaptionskip}{2pt}
\caption{KITTI ablation: binary rejection of semantic priors (ATE RMSE, m).}
\label{tab:ablation}
\begin{tabular*}{\columnwidth}{@{\extracolsep{\fill}}l|cccccc}
\noalign{\vskip 1pt}\hline\noalign{\vskip 1pt}
\textbf{Method} & 00 & 01 & 02 & 03 & 08 & 09 \\
\noalign{\vskip 1pt}\hline\noalign{\vskip 1pt}
ORB-SLAM3 & 1.27 & 13.11 & 5.68 & 1.30 & 3.72 & 1.87 \\
Hard Mismatch Rejection & 1.29 & 13.54 & 5.75 & 1.35 & 3.84 & 1.85 \\
Soft Filtering ($\tau_{\rm sem}=110$) & \textbf{1.25} & \textbf{12.08} & \textbf{5.47} & \textbf{1.29} & \textbf{3.54} & \textbf{1.84} \\
\noalign{\vskip 1pt}\hline
\end{tabular*}
\endgroup
\end{table}
\noindent\textbf{Backend optimisation budget.}
Because the priors are consumed in the frontend, a slower frontend could in
principle grant the mapping thread additional wall-clock, confounding the
comparison. We test this directly: a $50$\,ms per-frame delay inserted into
\emph{stock} ORB-SLAM3, where prior loading sits in our driver, raises
wall-clock $37$--$43\%$ and grants the mapping thread $12.9$--$58.3$\,s of
additional optimisation time across four sequences, yet ATE moves by at most
$0.02$\,m --- within the run-to-run variation of the three-run means reported
throughout (Supp.~D). Additional backend time confers no benefit, whereas the
modulation improves accuracy on every sequence at $6.8$--$8.2$\,ms/frame, a
seventh of the delay the baseline absorbed without effect.
\begin{table}[b]
\centering\small
\setlength{\tabcolsep}{5pt}
\begin{tabular}{@{}lcccc@{}}
\toprule
\textbf{Segmentation} & clean & 10\% & 30\% & 60\% \\
\cmidrule(r{4pt}){1-5}
Seq.~06 & 0.87 & 0.92 & 1.12 & 1.39 \\
\textit{\footnotesize keypoints affected} & -- & \footnotesize 0.9\% & \footnotesize 3.1\% & \footnotesize 6.4\% \\
Seq.~10 & 0.91 & 0.93 & 0.98 & 1.06 \\
\textit{\footnotesize keypoints affected} & -- & \footnotesize 0.5\% & \footnotesize 1.6\% & \footnotesize 3.1\% \\
\midrule
\textbf{Optical flow} & clean & 2.7\,px & 5.4\,px & 11.5\,px \\
\cmidrule(r{4pt}){1-5}
Seq.~06 & 0.87 & 0.87 & 1.01 & 1.00 \\
Seq.~10 & 0.91 & 0.91 & 1.00 & 1.10 \\
\bottomrule
\end{tabular}
\caption{Robustness to perception error, ATE RMSE (m), against ORB-SLAM3
baselines of $0.92$ and $1.32$\,m; \% of thing segments relabelled, mean flow
error on $44$--$48$\,px magnitudes. Keypoint exposure (Supp.~G) is the fraction
of detected keypoints receiving incorrect priors.}
\label{tab:robust}
\end{table}

\noindent\textbf{Modulation vs.\ rejection.}
Tab.~\ref{tab:ablation} isolates the core principle: hard rejection
\emph{degrades} accuracy on five of six sequences by thinning correspondence density, and soft filtering
yields only modest gains while remaining binary. Both contrast with the
consistent gains from continuous modulation (Tab.~\ref{tab:kitti},
\ref{tab:euroc}): contextual evidence works best by reshaping correspondence
cost, not pruning observations. Nor does any uniform reduction suffice: since
$\Phi_{\mathrm{pan}}$ scales $d$ below unity against ORB-SLAM3's absolute
thresholds, we scale every descriptor distance by $0.75\times$ in the stock
system with no priors, which degrades ATE on all sequences tested (Seq.~01,
04, 06: $13.11\!\to\!13.58$, $0.25\!\to\!0.26$, $0.92\!\to\!1.04$).

\noindent\textbf{Behaviour on static scenes.}
On EuRoC the motion module improves mean ATE reduction by $0.3$ points
($27.4\%\rightarrow27.7\%$) over the panoptic backbone and never regresses
below ORB-SLAM3. Feature survival under the stereo gate confirms the mechanism:
culling scales with scene difficulty, not object motion (Supp.~E). Since every
score is bounded to $[0,100]$, both gates operate on a plateau of the
flagged-\textit{dynamic} distribution rather than at a sensitive edge, so the
operating points transfer without per-scene adjustment (Supp.~F).
Qualitatively (Supp.~H), moving vehicles and pedestrians are flagged while
parked vehicles stay unpenalized within the same class: the per-segment
decision sets \emph{what} is penalized, the bounded score \emph{how much}.
\begin{table}[b]
\centering\small
\setlength{\tabcolsep}{3.5pt}
\begin{tabular}{@{}llccc@{}}
\toprule
Seq. & Method & $E_t$ (m) & $E_R$ ($^\circ$) & ATE (m) \\
\midrule
\multicolumn{5}{@{}l}{\textit{KITTI Tracking}~\cite{geiger2012kitti} (GPS/IMU ground truth)} \\
\multirow{2}{*}{0001} & ORB-SLAM3 & \textbf{0.0432} & 0.0357 & 2.277 \\
                      & KYS-SLAM  & \textbf{0.0432} & \textbf{0.0341} & \textbf{2.248} \\
\multirow{2}{*}{0018} & ORB-SLAM3 & 0.0440 & 0.0253 & 1.293 \\
                      & KYS-SLAM  & \textbf{0.0408} & \textbf{0.0250} & \textbf{1.140} \\
\midrule
\multicolumn{5}{@{}l}{\textit{Virtual KITTI~2}~\cite{cabon2020vkitti2} (exact ground truth)} \\
\multirow{2}{*}{Sc.~01} & ORB-SLAM3 & 0.0071 & 0.0193 & 0.240 \\
                        & KYS-SLAM  & \textbf{0.0066} & \textbf{0.0189} & \textbf{0.165} \\
\multirow{2}{*}{Sc.~18} & ORB-SLAM3 & 0.0109 & 0.0239 & 0.326 \\
                        & KYS-SLAM  & \textbf{0.0107} & \textbf{0.0239} & \textbf{0.282} \\
\multirow{2}{*}{Sc.~20} & ORB-SLAM3 & 0.0504 & \textbf{0.0259} & 11.720 \\
                        & KYS-SLAM  & \textbf{0.0473} & 0.0274 & \textbf{10.701} \\
\bottomrule
\end{tabular}
\vspace{-2mm}
\caption{Dynamic-scene evaluation under the same fixed configuration. $E_t$ and
$E_R$ are per-frame relative pose errors, following the convention of the
dynamic-SLAM literature; ATE is under $SE(3)$ alignment. Virtual KITTI~2
reproduces the KITTI Tracking sequences, so Sc.~01/0001 and Sc.~18/0018 are the
same trajectories under synthetic and real ground truth. Lower is better.}
\label{tab:dynamic}
\end{table}

\noindent\textbf{Robustness to perception error.}
Each learned input is perturbed with the others held fixed
(Tab.~\ref{tab:robust}). \emph{Segmentation:} a fraction of \emph{thing}
segments is reassigned to another observed class, with thing/stuff status
recomputed so misclassified objects enter the background ego-motion fit as a
genuine failure would. Accuracy degrades monotonically without tracking loss,
matching ORB-SLAM3 at ${\sim}10\%$ corruption on Seq.~06 and retaining the
advantage throughout on Seq.~10; the clean condition already contains the
segmenter's own errors, so the injected corruption is additional.
\emph{Optical flow:} zero-mean noise scaled by local magnitude gives errors
from $2.7$\,px --- already beyond current estimators --- to $11.5$\,px, yet
accuracy is unchanged at $2.7$\,px on both sequences. That tracking survives
every level tested follows from construction: the descriptor distance remains
primary and the priors only rescale it within bounded factors, so an incorrect
prior reweights a correspondence rather than removing it.

\noindent\textbf{Effect of the motion module.}
Comparing the backbone against the full model isolates the motion module's
contribution (Tab.~\ref{tab:kitti}). On the sequences with the most independent
motion (Seq.~02, 08, 10) the backbone reduces per-sequence ATE by
$\mathbf{17.7\%}$ and the motion module lifts this to $\mathbf{22.9\%}$,
against $14.3\%$ for Semantic~II and $8.3\%$ for RSO-SLAM. Dedicated dynamic
benchmarks (Tab.~\ref{tab:dynamic}) test the module where it matters most: mean
per-sequence ATE falls $6.6\%$ on KITTI Tracking and $17.8\%$ on Virtual
KITTI~2, reaching $31.2\%$ on Sc.~01. Sc.~18/0018 is the same trajectory under
exact synthetic and real GPS/IMU ground truth, and the gain holds under both
($13.5\%$, $11.8\%$); $E_R$ rises on Sc.~20. The module is not free where
motion is sparse or distant: it regresses appreciably against the backbone on
Seq.~00 and 01 while still improving on ORB-SLAM3 on all 21, and on EuRoC it
stays largely inactive. Seq.~00 contains little independent motion, and
Seq.~01 is the highway sequence, whose large depths and sparse near-field
structure degrade the disparity the ego-motion fit consumes. Disabling
continuous modulation while retaining the hard gates degrades ATE by
$11.7/4.1/23.1\%$ on Seq.~03, 09 and 10 (Supp.~B), so the contribution is not
gating alone.

\noindent\textbf{Detection limits.}
Detection is bounded by the scene's residual variability rather than by
threshold choice: distant slow movers fall within the background distribution,
while small distant segments yield noisier means and occasionally trip the
radial test on static structure. Neither is correctable by lowering thresholds,
as static vehicles can exhibit residuals exceeding those of true movers
(Supp.~I). Both errors are bounded --- an unflagged segment receives no
modulation, a false positive a score proportional to its weak evidence.
\vspace{-1mm}

\section{Conclusions and Future Work}
\label{sec:conclusion}
\vspace{-1mm}
KYS-SLAM is a modular stereo extension of ORB-SLAM3 fusing semantic, panoptic,
and zero-shot motion priors into one modulated descriptor distance, rescaling
correspondences by contextual plausibility rather than rejecting them. Under a
single fixed configuration it reduces per-sequence ATE by $17.4\%$ on KITTI and $27.7\%$ on
EuRoC, matching or improving all 21 sequences, and by 6.6\% and 17.8\% (up to
31.2\%) on dynamic subsets of KITTI Tracking and Virtual KITTI 2. Future work
targets online priors, learned features and end-to-end deployment.
{
    \small
    \bibliographystyle{ieeenat_fullname}
    \bibliography{main}
}

\clearpage
\appendix
\twocolumn[{%
\begin{center}
  \Large\bf Supplementary Material\\[12pt]
\end{center}
}]
\setcounter{table}{0}
\setcounter{figure}{0}
\renewcommand{\thetable}{S\arabic{table}}
\renewcommand{\thefigure}{S\arabic{figure}}
\renewcommand{\thesection}{\Alph{section}}
\setcounter{section}{0}

\section{Coefficient sensitivity}
\label{supp:panoptic}
Each coefficient is varied independently with the others held at their reported
values. Tab.~\ref{tab:supp_alpha} sweeps the geometric--semantic weighting
$\alpha$ of Eq.~(2) on KITTI Seq.~02, 07 and 08;
Tab.~\ref{tab:supp_eta} and Tab.~\ref{tab:supp_gamma} sweep the tier weights
$(\eta_1,\eta_2,\eta_3)$ and the same-instance discount $\gamma$ of Eq.~(5) on
Seq.~03, 09 and 10. The two stages consume different segmenters and therefore
different label spaces --- Eq.~(2) uses the ADE20K table of Sec.~3.1
($C\!=\!150$, range $[0,169]$) and Eq.~(5) the COCO panoptic table of Sec.~3.2
($133\times133$, range $[0,172]$, $S_{\max}\!=\!155$) --- so $\alpha$ and
$(\eta,\gamma)$ are not directly comparable across the two sweeps.

\noindent\textbf{Semantic weighting.} Accuracy is stable across the
geometry-dominant range ($\alpha\!\ge\!0.6$) and degrades sharply once the
semantic term dominates, reaching $10.40$\,m on Seq.~02 at $\alpha\!=\!0.1$.
The reported $\alpha\!=\!0.7$ lies in the interior of the stable range rather
than at its best point on every sequence: Seq.~08 is marginally better at
$\alpha\!=\!0.8$ ($3.40$ against $3.42$\,m), and Seq.~07 is insensitive over
the geometry-dominant range.

\noindent\textbf{Panoptic tiers.} Tab.~\ref{tab:supp_eta} sweeps the tier
weights by shifting mass off the geometric term $\eta_1$ ($0.90$ down to
$0.50$, a $1.8\times$ change) and onto the semantic and scale terms
$\eta_2\!+\!\eta_3$ ($0.10$ up to $0.50$, a fivefold change). Across that range
ATE varies by at most $27.5\%$ (Seq.~10, $1.16$ against $0.91$\,m) and by under
$15\%$ on the other two sequences ($10.7\%$ on Seq.~03 and $14.1\%$ on
Seq.~09), with a shallow interior minimum and no sharp optimum --- the
behaviour expected of scale-relative coefficients rather than fitted constants.
The reported setting is best on Seq.~03 and 10 and tied with
$(0.60,0.25,0.15)$ on Seq.~09.

Both endpoints of the $\gamma$ sweep (Tab.~\ref{tab:supp_gamma}) degrade
accuracy on all three sequences: over-discounting ($\gamma=0.60$) costs
$10.0$--$20.9\%$ and removing the discount entirely ($\gamma=1.00$, i.e.\ no
same-instance preference) costs $2.4$--$30.8\%$. The reported $\gamma=0.75$ is
best on Seq.~09 and 10 and tied with $\gamma=0.70$ on Seq.~03. The
same-instance tier therefore performs measurable work rather than serving as an
unmotivated refinement.

\begin{table}[h]
\centering\small
\setlength{\tabcolsep}{3.5pt}
\begin{tabular}{@{}lcccccccc@{}}
\toprule
$\alpha$ & 1.0 & 0.9 & 0.8 & \textbf{0.7} & 0.6 & 0.5 & 0.3 & 0.1 \\
\midrule
Seq.~02 & 5.68 & 5.48 & 4.81 & \textbf{4.66} & 5.65 & 5.62 & 4.99 & 10.40 \\
Seq.~07 & 0.47 & 0.48 & 0.47 & \textbf{0.47} & 0.52 & 0.50 & 0.76 & 0.53 \\
Seq.~08 & 3.72 & 3.48 & 3.40 & \textbf{3.42} & 3.80 & 3.73 & 3.60 & 3.75 \\
\bottomrule
\end{tabular}
\caption{Geometric--semantic weighting $\alpha$, ATE RMSE (m). Bold marks the
reported configuration, not the per-sequence best; settings that match or beat
it are identified in the text. $\alpha=1.0$ reduces Eq.~(2) to the raw ORB
distance and recovers stock ORB-SLAM3.}
\label{tab:supp_alpha}
\end{table}

\begin{table}[h]
\centering\small
\setlength{\tabcolsep}{4pt}
\begin{tabular}{@{}cccccc@{}}
\toprule
$\eta_1$ & $\eta_2$ & $\eta_3$ & Seq.~03 & Seq.~09 & Seq.~10 \\
\midrule
0.90 & 0.05 & 0.05 & 1.14 & 1.94 & 1.16 \\
0.80 & 0.15 & 0.05 & 1.05 & 1.84 & 1.13 \\
\textbf{0.70} & \textbf{0.20} & \textbf{0.10} & \textbf{1.03} & \textbf{1.70} & \textbf{0.91} \\
0.60 & 0.25 & 0.15 & 1.09 & 1.70 & 0.98 \\
0.50 & 0.30 & 0.20 & 1.11 & 1.80 & 1.10 \\
\bottomrule
\end{tabular}
\caption{Panoptic tier weights, ATE RMSE (m). Bold marks the reported
configuration, not the per-sequence best; $(0.60,0.25,0.15)$ ties it on
Seq.~09.}
\label{tab:supp_eta}
\end{table}

\begin{table}[h]
\centering\small
\setlength{\tabcolsep}{6pt}
\begin{tabular}{@{}cccc@{}}
\toprule
$\gamma$ & Seq.~03 & Seq.~09 & Seq.~10 \\
\midrule
0.60 & 1.17 & 1.87 & 1.10 \\
0.70 & 1.03 & 1.72 & 1.04 \\
\textbf{0.75} & \textbf{1.03} & \textbf{1.70} & \textbf{0.91} \\
0.85 & 1.08 & 1.71 & 1.05 \\
1.00 & 1.20 & 1.74 & 1.19 \\
\bottomrule
\end{tabular}
\caption{Same-instance discount $\gamma$, ATE RMSE (m). Bold marks the reported
configuration, not the per-sequence best; $\gamma=0.70$ ties it on Seq.~03.
$\gamma=1.00$ removes the same-instance preference; with $\eta_1$ held fixed
this also inverts the tier ordering, so the degradation bounds the tier's
contribution rather than isolating it.}
\label{tab:supp_gamma}
\end{table}

\section{Motion sensitivity}
\label{supp:motion}
The motion sensitivity triple $\lambda_t$ of Eq.~(9) is scaled by a common
factor, preserving the tier ordering justified in Sec.~3.4 of the main paper
and varying only its magnitude (Tab.~\ref{tab:supp_lambda}). Accuracy is flat
between $0.5\times$ and $2\times$, and degrades at $4\times$ on Seq.~09 and 10,
where over-penalisation begins discarding usable correspondences. The reported
$1\times$ setting is best on Seq.~03 and 10; $2\times$ is marginally better on
Seq.~09 ($1.67$ against $1.70$\,m) and marginally worse on Seq.~10 ($0.92$
against $0.91$\,m).

The $0\times$ configuration retains the hard motion gates of Sec.~3.4 while
disabling continuous modulation, isolating the graded term specifically. It
degrades ATE by $11.7\%$, $4.1\%$ and $23.1\%$ on Seq.~03, 09 and 10 relative
to the reported setting, so the motion module's contribution is not reducible
to its gating.

The comparison is sharper against the panoptic backbone alone, which omits the
motion module entirely (Tab.~1 of the main paper: $1.14$, $1.75$ and
$0.99$\,m). Gating without modulation is no better than having no motion
module at all --- $0.9\%$ and $1.1\%$ worse on Seq.~03 and 09 and $13.1\%$
worse on Seq.~10 --- so the hard gates recover none of the module's benefit on
their own. They are retained because they alone remove correspondences whose
motion evidence is extreme enough that no descriptor similarity remains
meaningful (Sec.~3.4 of the main paper), a regime the bounded multiplicative
penalty of Eq.~(9) cannot reach by construction. The reported configuration
uses both, and only their combination improves on the backbone.

\begin{table}[h]
\centering\small
\setlength{\tabcolsep}{4pt}
\begin{tabular}{@{}lcccc@{}}
\toprule
Scale & $\lambda_t$ & Seq.~03 & Seq.~09 & Seq.~10 \\
\midrule
$0\times$   & (0, 0, 0)           & 1.15 & 1.77 & 1.12 \\
$0.5\times$ & (0.04, 0.025, 0.01) & 1.06 & 1.76 & 0.99 \\
$\mathbf{1\times}$ & \textbf{(0.08, 0.05, 0.02)} & \textbf{1.03} & \textbf{1.70} & \textbf{0.91} \\
$2\times$   & (0.16, 0.10, 0.04)  & 1.07 & 1.67 & 0.92 \\
$4\times$   & (0.32, 0.20, 0.08)  & 1.07 & 1.84 & 1.12 \\
\bottomrule
\end{tabular}
\caption{Motion sensitivity, ATE RMSE (m). Bold marks the reported
configuration, not the per-sequence best. The $0\times$ row retains the hard
gates while disabling continuous modulation; the panoptic backbone without the
motion module scores $1.14$, $1.75$ and $0.99$\,m on these sequences (Tab.~1 of
the main paper).}
\label{tab:supp_lambda}
\end{table}

\section{Runtime and system cost}
\label{supp:runtime}
Measurements were taken on an NVIDIA RTX~A6000 (48\,GB, driver 560.35.05, CUDA
12.6) with the core clock pinned to eliminate thermal variance. Prior
generation (Tab.~\ref{tab:supp_stage}) is averaged over 300 consecutive frames
of KITTI~06 across three runs; the online costs of
Tab.~\ref{tab:supp_online} are measured under the same protocol on Seq.~03, 04
and 06, over 300 consecutive frames or the full sequence where shorter
(Seq.~04, 271 frames). Per-frame costs exclude one-time process initialisation,
which is $\approx4.8$\,s, consistent across components and varying by under
$0.6$\,s between runs.

Tab.~\ref{tab:supp_stage} decomposes prior generation. Optical flow and
panoptic segmentation are run on the left and right images separately, so each
contributes twice to the per-frame sum; stereo disparity is computed once from
the pair. The per-pass costs are measured on the left image and doubled;
independently timed right-image passes agree to within $2$--$3$\,ms, as
expected for fixed-resolution inference. Panoptic segmentation comprises
$266$\,ms inference and $74$\,ms postprocess, a split that agrees to within
$1.1$\,ms between two independent runs. Disparity and the two flow passes are
mutually independent and parallelise: run concurrently on a single GPU these
three stages take $840$\,ms/frame against $1016$\,ms in sequence, a
$1.21\times$ gain indicating the GPU is already well utilised.

\begin{table}[h]
\centering\small
\setlength{\tabcolsep}{5pt}
\begin{tabular}{@{}lr@{}}
\toprule
Stage & Cost \\
\midrule
Stereo disparity                     & 214 \\
Optical flow (left, right)           & $2\times401$ \\
Panoptic segmentation (left, right)  & $2\times340$ \\
\midrule
\textbf{GPU sequential sum}          & \textbf{1696} \\
CPU-side fusion                      & 23.4 \\
\midrule
\textbf{Total}                       & \textbf{1719.4} \\
\bottomrule
\end{tabular}
\caption{Per-frame prior generation cost (ms). Optical flow and panoptic
segmentation run once per camera; disparity is computed once from the stereo
pair. The three GPU stages are off-the-shelf perception models; the
$23.4$\,ms of CPU-side fusion is introduced here.}
\label{tab:supp_stage}
\end{table}

Tab.~\ref{tab:supp_online} reports the online cost against stock ORB-SLAM3. The
modulation row is the difference between the two tracking rows above it. Prior
deserialisation is timed outside the tracking window and reported separately,
so the tracking delta isolates the modulation itself.

\noindent\textbf{What is and is not real-time.} With priors resident,
KYS-SLAM tracking runs at $32.6$--$35.7$\,ms/frame ($28$--$31$\,Hz) and, with
deserialisation included, at $53.2$--$55.4$\,ms/frame ($18$--$19$\,Hz) --- both
above the $10$\,Hz capture rate of KITTI. The matching stage introduced here is
therefore real-time on this hardware; what is not is the off-the-shelf
perception front-end of Tab.~\ref{tab:supp_stage}, which is precomputed offline
in all experiments, as in prior semantic-SLAM systems. Accounting separates
cleanly along the same line: everything specific to KYS-SLAM --- modulation,
CPU-side fusion, and deserialisation --- sums to $49.2$--$51.5$\,ms/frame, or
$2.8$--$2.9\%$ of the full per-frame pipeline, of which the modulation itself
is $\approx0.4\%$ ($0.38$--$0.46\%$ across the three sequences). The
$19$--$21$\,ms of deserialisation is an artefact of file-based prior storage
rather than of the formulation.

\noindent\textbf{Memory.} Peak host memory rises by $0.22$--$0.52$\,GB, which
is $32$--$43\%$ above the stock peak RSS and is dominated by the resident prior
buffers; unlike the time cost, it is not a small fraction of the baseline and
we do not present it as one.

\begin{table}[h]
\centering\small
\setlength{\tabcolsep}{4pt}
\begin{tabular}{@{}lccc@{}}
\toprule
& Seq.~03 & Seq.~04 & Seq.~06 \\
\midrule
ORB-SLAM3 tracking (ms)   & 25.1 & 27.5 & 27.9 \\
KYS-SLAM tracking (ms)    & 32.6 & 35.7 & 34.7 \\
\textbf{Modulation (ms)}  & \textbf{7.5} & \textbf{8.2} & \textbf{6.8} \\
Prior deserialisation (ms)& 20.6 & 19.7 & 19.0 \\
\midrule
ORB-SLAM3 peak RSS (GB)   & 0.80 & 0.68 & 1.34 \\
KYS-SLAM peak RSS (GB)    & 1.14 & 0.90 & 1.86 \\
$\Delta$ peak RSS (GB)    & +0.34 & +0.22 & +0.52 \\
\bottomrule
\end{tabular}
\caption{Online cost against stock ORB-SLAM3, mean per-frame tracking time and
peak host memory. Prior deserialisation is measured outside the tracking
window, so the modulation row isolates the matching cost.}
\label{tab:supp_online}
\end{table}

\section{Backend optimisation budget}
\label{supp:backend}
To test whether a slower frontend confounds the comparison by granting the
mapping thread additional wall-clock, an artificial $50$\,ms per-frame delay
was inserted into \emph{stock} ORB-SLAM3, positioned where prior loading sits
in our driver and excluded from the reported tracking time. Four sequences were
re-run under the standard sleep-to-timestamp pacing
(Tab.~\ref{tab:supp_stall}).

The delay is not absorbed by the frame budget. The mapping thread receives
$94$--$99\%$ of the injected budget as real additional optimisation time
($\Delta t$ against $50$\,ms $\times$ frame count), so wall-clock rises
$37$--$43\%$ and the thread gains between $12.9$\,s and $58.3$\,s, a
$4.5\times$ range across sequence lengths from 271 to 1201 frames. ATE changes
by at most $0.02$\,m, in inconsistent directions, and within the run-to-run
variation of the three-run means reported in the main paper. Additional
backend time therefore confers no measurable benefit, and the
$6.8$--$8.2$\,ms/frame the modulation costs --- roughly a seventh of the delay
absorbed here without effect --- cannot account for the accuracy gain.

\begin{table}[h]
\centering\small
\setlength{\tabcolsep}{3.5pt}
\begin{tabular}{@{}lccccc@{}}
\toprule
Seq. & Frames & Base (s) & +50\,ms (s) & $\Delta t$ (s) & ATE (m) \\
\midrule
03 & 801  & 94.4  & 132.0 & +37.6 & 1.30 $\to$ 1.31 \\
04 & 271  & 34.6  & 47.5  & +12.9 & 0.25 $\to$ 0.27 \\
06 & 1101 & 126.0 & 180.6 & +54.6 & 0.92 $\to$ 0.91 \\
10 & 1201 & 138.1 & 196.4 & +58.3 & 1.32 $\to$ 1.32 \\
\bottomrule
\end{tabular}
\caption{Backend optimisation budget control. $\Delta t$ is the additional
wall-clock granted to the mapping thread. ATE is that of stock ORB-SLAM3
without and with the injected delay.}
\label{tab:supp_stall}
\end{table}

\section{Feature survival under the stereo gate}
\label{supp:survival}

\begin{figure}[h]
  \centering
  \includegraphics[width=\linewidth,height=0.25\textheight,keepaspectratio]{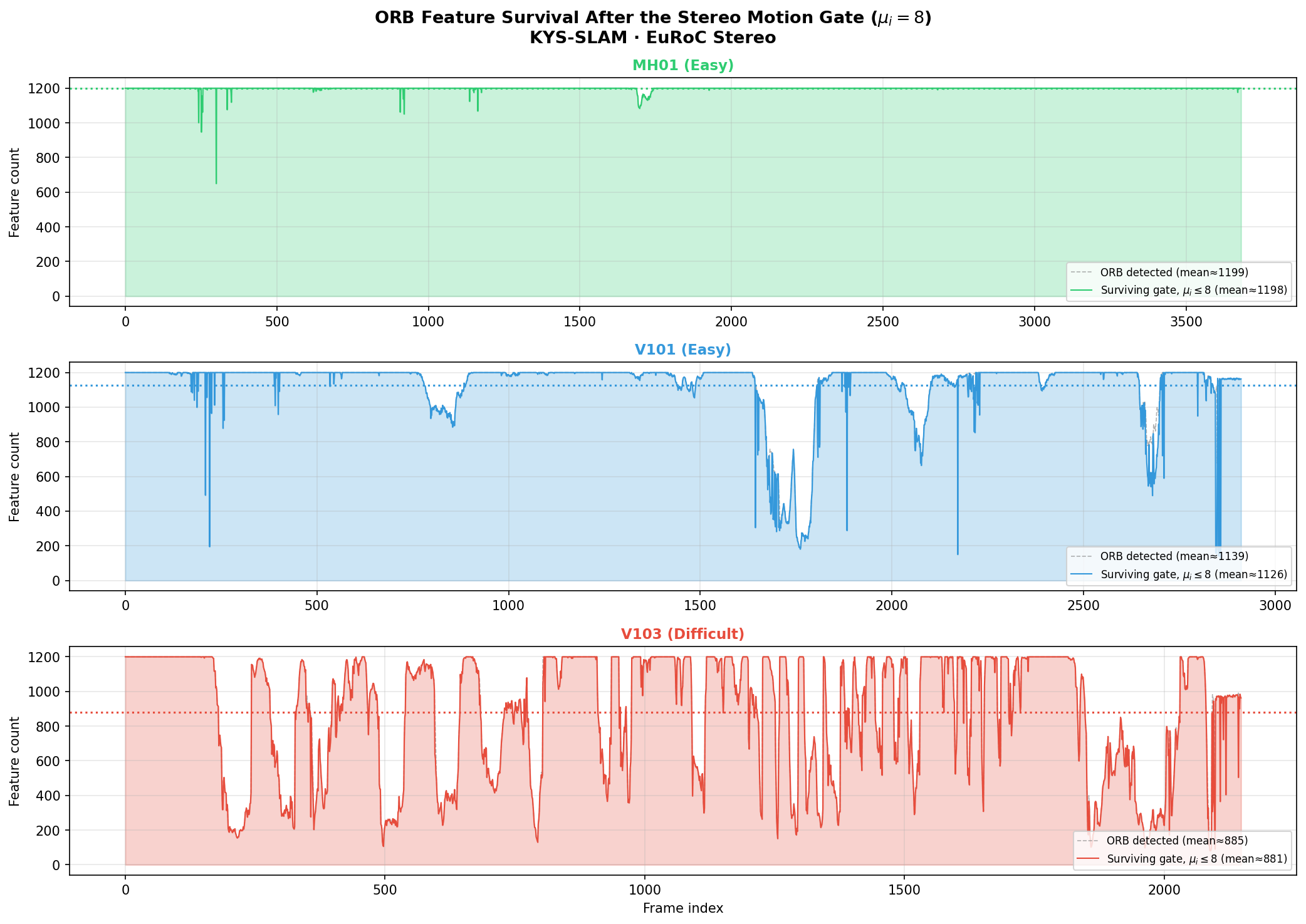}
  \vspace{-3mm}
  \caption{ORB feature survival after the stereo motion gate ($\mu_i{=}8$) on
  EuRoC, against the 1200 features extracted per frame. Culling scales with
  scene difficulty: nearly inert on static MH01 ($\approx1198$, $99.8\%$),
  modest on V101 ($\approx1126$, $93.8\%$), larger on V103 ($\approx881$,
  $73.4\%$), consistent with the rapid rotation and motion blur characterising
  V103.}
  \label{fig:feature_survival}
  \vspace{-3mm}
\end{figure}

Fig.~\ref{fig:feature_survival} reports ORB feature survival after the stereo
motion gate on three EuRoC sequences of increasing difficulty. Culling tracks
scene difficulty rather than object motion, driven by rapid rotation and motion
blur on the harder sequences.

V103 is the relevant stress case for both conditions: it combines rapid
rotation with pronounced motion blur, and culling there is heaviest
($\approx73\%$ survival). Tracking is nonetheless maintained and accuracy still
improves on ORB-SLAM3 ($0.050$ against $0.061$\,m, Tab.~2 of the main paper),
which follows from construction --- the gate is unilateral and bounded, and
keypoints without valid panoptic labels bypass modulation entirely rather than
being discarded.

\section{Motion-score distribution}
\label{supp:distribution}

\begin{figure}[h]
  \centering
  \includegraphics[width=\linewidth,height=0.22\textheight,keepaspectratio]{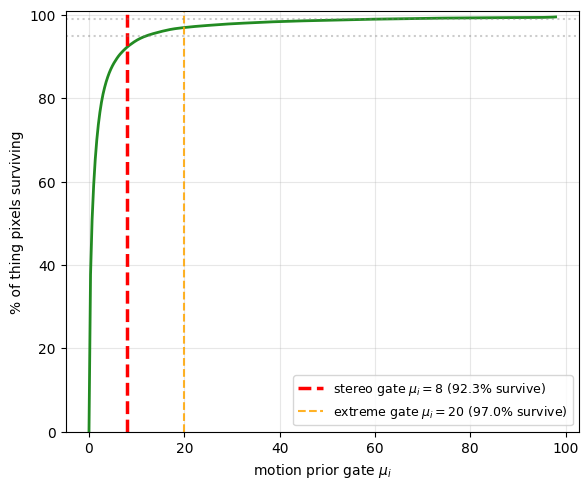}
  \vspace{-3mm}
  \caption{Motion-score survival on KITTI Seq.~08, over pixels of
  \textit{thing} segments flagged \textit{dynamic} by the detection module
  (static \textit{stuff} and unflagged \textit{things} are gated to zero
  upstream and excluded). The curve plateaus after the initial rise, placing
  the stereo gate ($\mu_i{=}8$, retaining $92.3\%$) and the extreme gate
  ($\mu_i{=}20$, retaining $97.0\%$) on the same flat region.}
  \label{fig:distribution}
  \vspace{-3mm}
\end{figure}

Fig.~\ref{fig:distribution} shows the distribution of the exported motion prior
over pixels of \textit{thing} segments flagged dynamic. Because every score is
bounded to $[0,100]$, both gates fall on the plateau of that distribution
rather than at a sensitive edge, so the operating points transfer across
sequences without per-scene adjustment. The constant of Eq.~(8) and the two
gate positions are consequently one degree of freedom, not three: the constant
sets only the scale on which the gates are read.

\section{Robustness to perception error}
\label{supp:robustness}
Tab.~4 \emph{of the main paper} reports the results; we detail the perturbation
design and the exposure mechanism here.

\noindent\textbf{Segmentation.} A fraction of \textit{thing} segments is
reassigned to another class observed in the sequence, with thing/stuff status
recomputed so that misclassified objects are admitted into the background
ego-motion fit as a genuine failure would admit them. Segment geometry,
disparity and optical flow are held fixed, so panoptic labelling is the only
variable. Corruption is nested: segments hit at $10\%$ are a subset of those
hit at $30\%$ and $60\%$, so each level is strictly more perturbed than the
last. The clean condition already contains the segmenter's own errors, so the
injected corruption is additional rather than total.

The perturbation exceeds what the segment fraction implies. Because
misclassified objects are textured and proximal, they attract ORB features
disproportionately: keypoint-level exposure, measured with the ORB extractor
settings used by ORB-SLAM3 on KITTI, is amplified $1.6$--$1.9\times$ relative
to pixel area on both sequences. Exposure scales with dynamic content and
accounts for the difference in degradation between the two sequences; on
Seq.~06 at $60\%$ corruption it reaches $13.1\%$ of keypoints at the $95$th
percentile and $43.7\%$ in the most affected frame.

Degradation is monotonic and tracking is never lost. Seq.~10 stays below the
ORB-SLAM3 baseline ($1.32$\,m) at every corruption level; Seq.~06 matches its
baseline ($0.92$\,m) exactly at $10\%$ corruption and concedes above it
thereafter.

\noindent\textbf{Optical flow.} Zero-mean noise scaled by local flow magnitude
is injected into the estimated field, with labels, geometry and disparity held
fixed. Measured against the unperturbed field this corresponds to mean errors
of $2.7$, $5.4$ and $11.5$\,px on flow magnitudes averaging $44$--$48$\,px ---
the lowest setting already exceeding the accuracy of current estimators on
KITTI, and roughly an order of magnitude worse at the highest. Seq.~10 remains
below the ORB-SLAM3 baseline ($1.32$\,m) at every level; Seq.~06 ($0.92$\,m) at
$2.7$\,px, conceding at most $0.09$\,m above it. We note the perturbation is
per-pixel independent, whereas real flow failures are spatially correlated at
occlusion boundaries and on textureless surfaces; the reported tolerance
therefore characterises sensitivity to estimator precision rather than to
structured failure.

\section{Qualitative motion classification}
\label{supp:qualitative}

\begin{figure}[h]
  \centering
  \includegraphics[width=\linewidth]{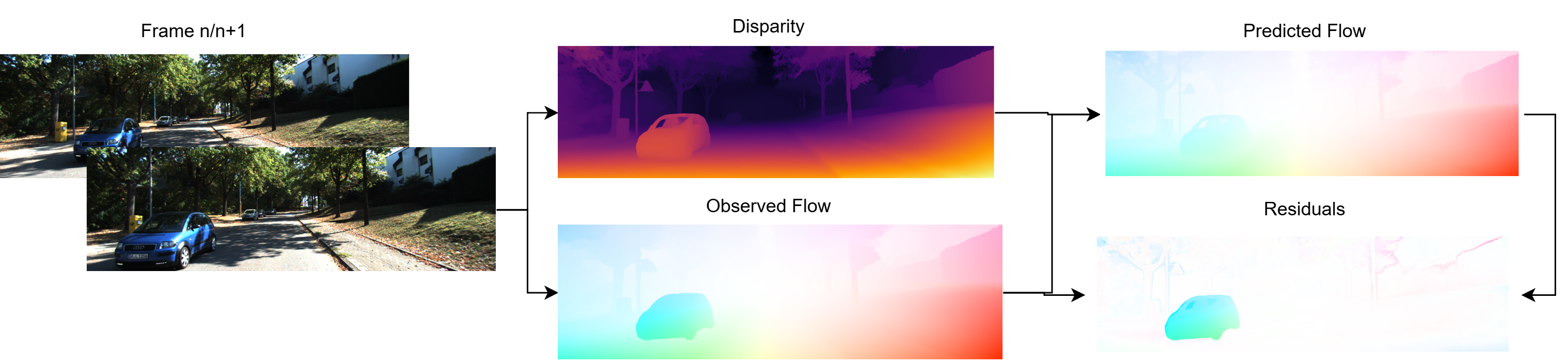}
  \vspace{-2mm}
  \caption{Raw residual extraction: observed flow $\mathbf{F}_{\mathrm{obs}}$
  and disparity-derived depth give the predicted ego-motion flow
  $\mathbf{F}_{\mathrm{m}}$, whose difference is the residual
  $\mathbf{r}=\mathbf{F}_{\mathrm{obs}}-\mathbf{F}_{\mathrm{m}}$.}
  \label{fig:residual_extraction}
  \vspace{-2mm}
\end{figure}

\begin{figure}[h]
    \centering
    \includegraphics[width=\linewidth,height=0.30\textheight,keepaspectratio]{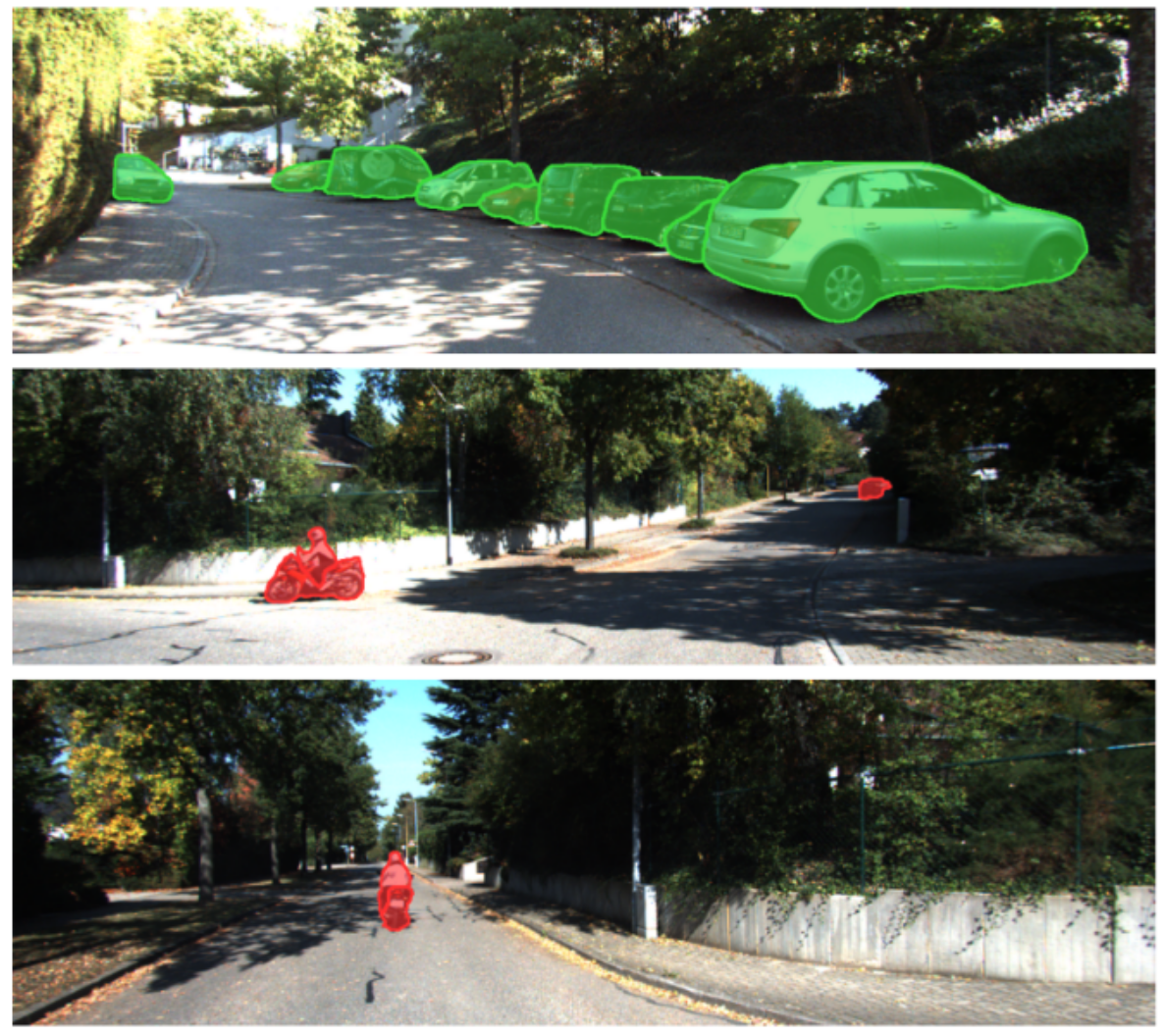}
    \vspace{-1mm}
    \includegraphics[width=\linewidth,height=0.30\textheight,keepaspectratio]{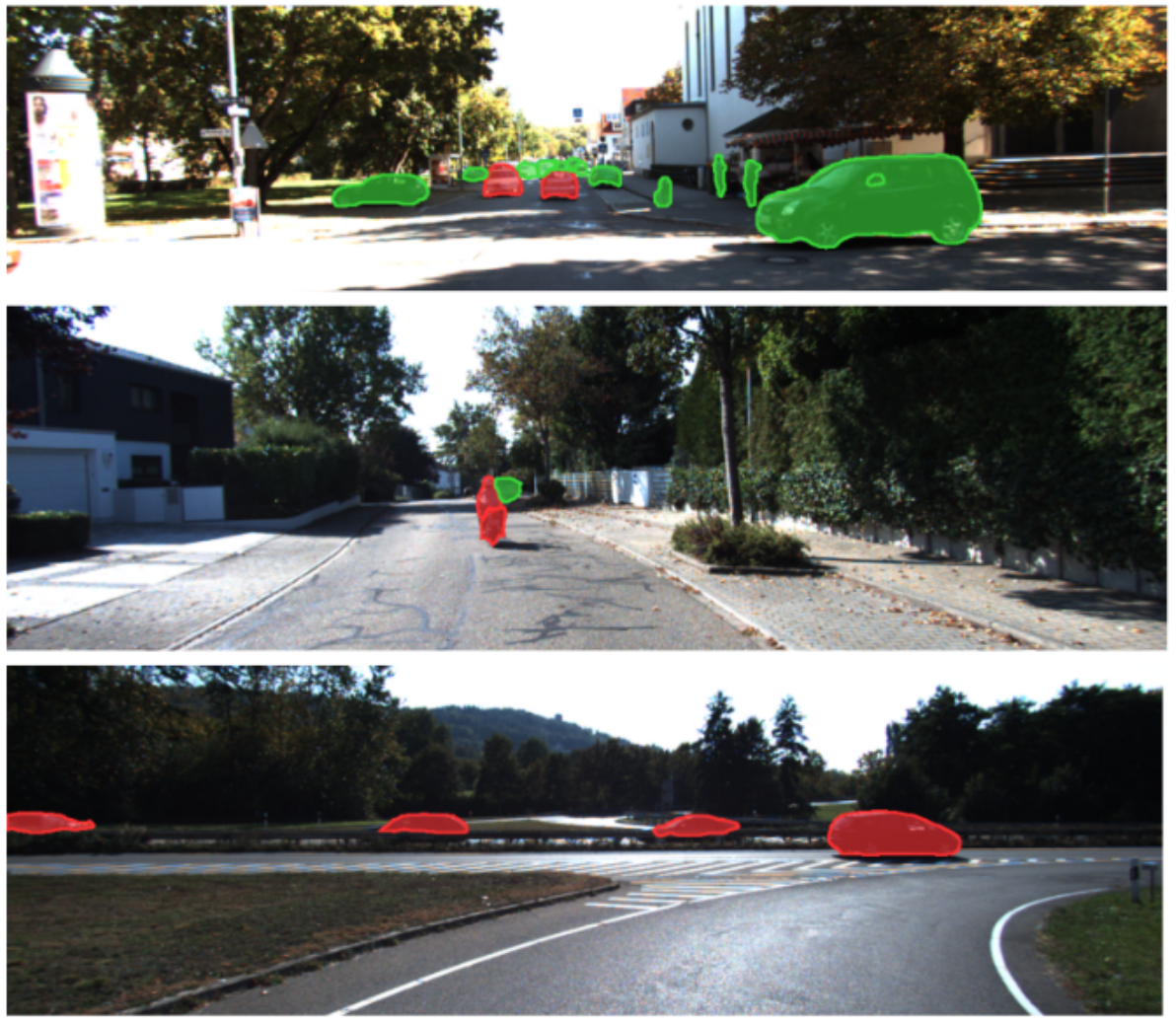}

    \caption{Qualitative motion classification from the proposed zero-shot
    motion model. Static regions are shown in green and independently dynamic
    regions in red. Parked vehicles and static structure stay unpenalized even
    within the same class as flagged movers.}
    \label{fig:motion_qualitative}
    \vspace{-3mm}
\end{figure}

Fig.~\ref{fig:residual_extraction} illustrates the residual extraction of
Sec.~3.3 of the main paper, and Fig.~\ref{fig:motion_qualitative} the resulting
per-segment classification: moving vehicles and pedestrians are flagged while
parked vehicles stay unpenalized within the same class, so the per-segment
decision sets what is penalized and the bounded score how much.

\section{Detection limits}
\label{supp:limits}

Detection is bounded by the scene's own residual variability rather than by
threshold choice, and both error directions are bounded in consequence.

\begin{figure*}
    \centering
    \includegraphics[width=1\linewidth]{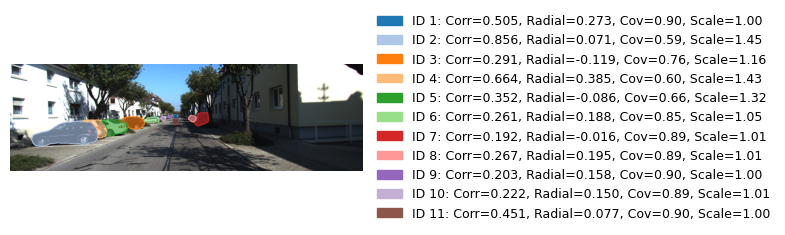}
    \includegraphics[width=1\linewidth]{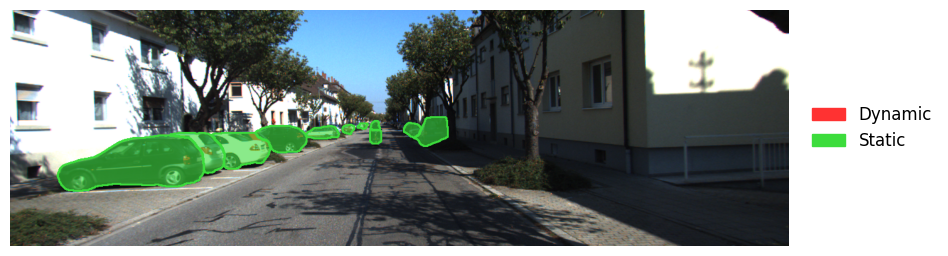}
  \caption{Missed mover. A motorcyclist, segmented as two entities (rider,
  ID~1; vehicle, ID~9), is not flagged: corrected residuals and radial
  components fall within the per-frame background distribution, against
  thresholds of $2.23$ and $\pm1.75$.}
  \label{fig:supp_missed}
\end{figure*}

\noindent\textbf{Missed movers.} A motorcyclist, segmented as two entities
(rider and vehicle), is not flagged (Fig.~\ref{fig:supp_missed}): corrected
residuals of $0.51$ and $0.20$ and radial components of $0.27$ and $0.16$ fall
within the background distribution for that frame ($1.10\pm0.56$ and
$-0.15\pm0.88$), against thresholds of $2.23$ and $\pm1.75$. Three difficulties
compound --- the object is distant and small, so its projected motion
approaches the residual noise floor; it is thin and self-occluding against
textureless road; and panoptic segmentation separates rider from vehicle, so
neither segment aggregates the full motion evidence.

\begin{figure*}
    \centering
    \includegraphics[width=1\linewidth]{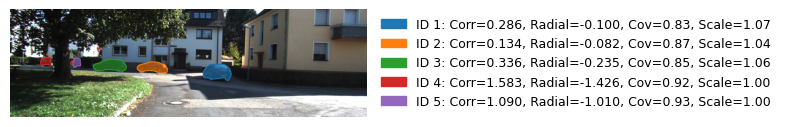}
    \includegraphics[width=\linewidth]{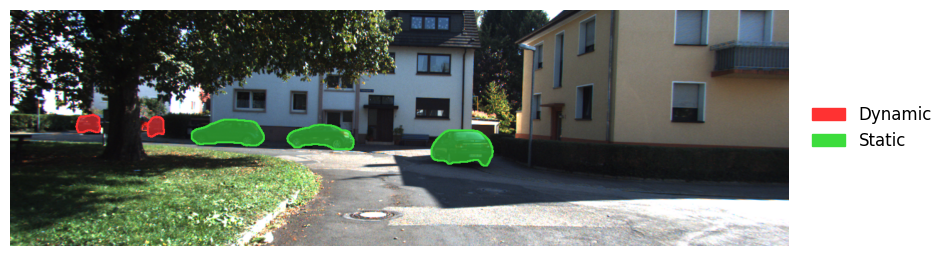}
     \caption{False positives. Two parked vehicles (IDs~4 and 5, at $20.1$ and
  $17.8$\,m) are flagged by the radial test alone. Background coverage is high
  ($0.92$, $0.93$), so no threshold inflation applies; the distinguishing
  factor is segment size.}
    \label{fig:supp_fp}
\end{figure*}

\noindent\textbf{False positives.} Two parked vehicles at $20.1$ and $17.8$\,m
are flagged by the radial test alone (Fig.~\ref{fig:supp_fp}); their corrected
magnitudes ($1.58$, $1.09$) remain below the magnitude threshold of $2.90$, and
background coverage is high ($0.92$, $0.93$) so no threshold inflation applies.
Depth is not the cause: their depth spreads ($1.08$, $1.84$\,m) are comparable
to that of a correctly classified vehicle ($1.47$\,m). The distinguishing
factor is segment size --- $907$ and $693$\,px against $3079$--$4380$\,px for
the correctly classified vehicles. The per-segment statistic is a mean over the
segment mask, so its sampling error grows as segments shrink, while the
threshold is two standard deviations of a background distribution estimated
over the full image and unusually tight on this straight trajectory
($\sigma=0.174$ against $\sigma=0.966$ for the corrected magnitude).

Neither error is correctable by lowering thresholds: in the missed-mover frame
of Fig.~\ref{fig:supp_missed}, two static vehicles exhibit residuals ($0.856$,
$0.664$) exceeding those of the true mover, so any threshold admitting the
latter would flag the former first. Consequences are bounded by construction:
an unflagged segment receives no modulation, and a false positive carries a
score proportional to its weak evidence, so the correspondence is attenuated by
a small fraction of the descriptor distance rather than discarded. A binary
detector making either error would remove the affected features outright.

Taken with the segmentation and flow perturbations of Sec.~\ref{supp:robustness}
and the rotation- and blur-dominated sequences of Sec.~\ref{supp:survival},
these two cases cover the failure modes of the motion module: incorrect
panoptic labels, degraded optical flow, rapid camera rotation, motion blur, and
small or distant dynamic objects. In each, an incorrect prior rescales a
correspondence within bounded factors rather than removing it, which is why
accuracy degrades gradually and tracking survives every condition tested.

\section{DROID-SLAM baseline protocol}
\label{supp:droid}

The original DROID-SLAM paper reports stereo EuRoC results in its supplementary
material, obtained with their network trained on synthetic monocular video, and
states that in the stereo setting the camera trajectory is recovered ``up to
scale''. We therefore report DROID-SLAM$^{*}$, the baseline retrained from
scratch for stereo by DVI-SLAM under a controlled setting, unchanged from the
published table (Sec.~4).

\end{document}